\documentclass[final]{cvpr}

\usepackage{times}
\usepackage{epsfig}
\usepackage{graphicx}
\usepackage[dvipsnames]{xcolor}
\usepackage{mathtools}
\usepackage{amssymb}
\usepackage{booktabs}
\usepackage{caption}
\usepackage{subcaption}
\usepackage{multirow}
\usepackage{amsthm}
\usepackage{lipsum}
\usepackage{mwe}
\usepackage{comment}
\usepackage{cite}
\usepackage{float}
\usepackage[export]{adjustbox}

\usepackage{multicol}
\newcommand\blfootnote[1]{%
  \begingroup
  \renewcommand\thefootnote{}\footnote{#1}%
  \addtocounter{footnote}{-1}%
  \endgroup
}

\usepackage{hyperref}
\hypersetup{
    colorlinks=true,
    linkcolor=blue,
    filecolor=magenta,      
    urlcolor=cyan,
}

\usepackage{cleveref}

\def\cvprPaperID{****} 
\def\confYear{CVPR 2021}

\def\im{}
\def\loss{\mathcal{L}}
\def\feat{\mathcal{F}}
\def\gram{\mathcal{G}}
\def\range{\mathcal{V}}

\DeclareMathOperator{\mse}{MSE}

\def\ist{IST\,} 

\begin{document}

\title{Style-Aware Normalized Loss for Improving Arbitrary Style Transfer}
\author{Jiaxin Cheng\textsuperscript{*\dag}, Ayush Jaiswal\textsuperscript{\ddag}, Yue Wu\textsuperscript{\ddag}, Pradeep Natarajan\textsuperscript{\ddag}, Prem Natarajan\textsuperscript{\ddag}\\
\textsuperscript{\dag}USC Information Sciences Institute, \textsuperscript{\ddag}Amazon Alexa Natural Understanding\\
{\tt\small chengjia@isi.edu, \{ayujaisw,wuayue,natarap,premknat\}@amazon.com}
}

\twocolumn[{%
\renewcommand\twocolumn[1][]{#1}%
\maketitle
\vspace{-25pt}
\begin{center}
    \centering
    \includegraphics[width=\textwidth,trim=0.0cm 0.5cm 0.0cm 0.15cm,clip]{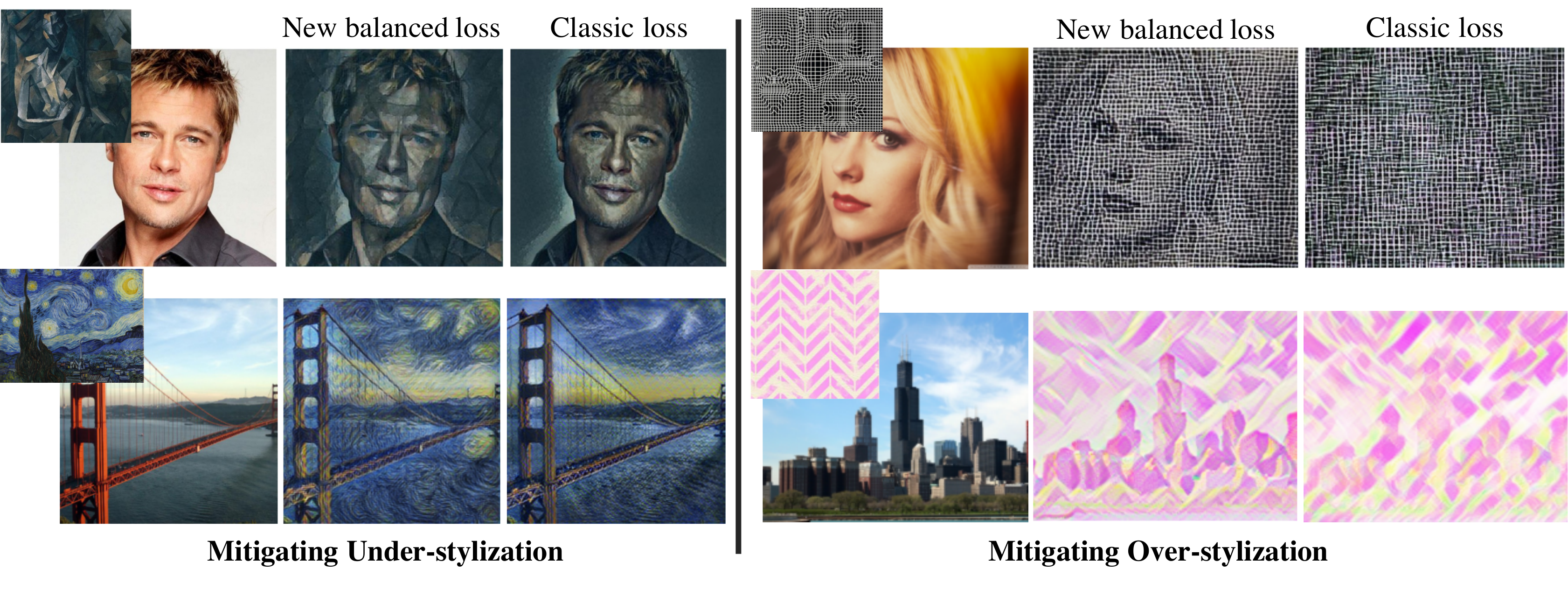}
    \captionof{figure}{\label{fig.teaser} Popular Arbitrary Style Transfer methods suffer from under-stylization and over-stylization due to imbalanced style transferability during training. Our new balanced loss mitigates these under- and over-stylization issues.}\label{fig.teaser}
\end{center}%
}]

\pagenumbering{gobble}

\begin{abstract}
Neural Style Transfer (NST)\blfootnote{\textsuperscript{*}This work was completed during his internship at Amazon.} has quickly evolved from single-style to infinite-style models, also known as Arbitrary Style Transfer (AST). Although appealing results have been widely reported in literature, our empirical studies on four well-known AST approaches (GoogleMagenta\cite{megenta2}, AdaIN\cite{AdaIn}, LinearTransfer\cite{LinearTransfer}, and SANet\cite{SANet}) show that more than 50\% of the time, AST stylized images are not acceptable to human users, typically due to under- or over-stylization. We systematically study the cause of this imbalanced style transferability (\ist) and propose a simple yet effective solution to mitigate this issue. Our studies show that the \ist issue is related to the conventional AST style loss, and reveal that the root cause is the equal weightage of training samples irrespective of the properties of their corresponding style images, which biases the model towards certain styles. Through investigation of the theoretical bounds of the AST style loss, we propose a new loss that largely overcomes \ist. Theoretical analysis and experimental results validate the effectiveness of our loss, with over 80\% relative improvement in style deception rate and 98\% relatively higher preference in human evaluation.
\vspace{-5pt}
\end{abstract}

\vspace{-35pt}
\section{Introduction}
Neural style transfer (NST) refers to the generation of a pastiche image $\im{P}$ from two images $\im{C}$ and $\im{S}$ via a neural network, where $\im{P}$ shares the content with $\im{C}$ but is in the style of $\im{S}$. While the original NST approach of Gatys\cite{gatys2016image} optimizes the transfer model for each pair of $\im{C}$ and $\im{S}$, the field has rapidly evolved in recent years to develop models that support arbitrary styles out-of-the-box. NST models can, hence, be classified based on their stylization capacity into models trained for (1) a single combination of $\im{C}$ and $\im{S}$ \cite{gatys2016image,li2017demystifying,li2017laplacian,risser2017stable,kolkin2019style}, (2) one $\im{S}$ \cite{johnson2016perceptual,ulyanov2016texture,ulyanov2017improved,li2016precomputed}, (3) multiple fixed $\im{S}$ \cite{chen2017stylebank,dumoulin2016learned,li2017diversified,zhang2018multi,kotovenko2019content2,sanakoyeu2018style}, and (4) infinite (arbitrary) $\im{S}$ \cite{WCT,AdaIn,megenta2,SANet,LinearTransfer,chen2016fast,jing2020dynamic,gu2018arbitrary,shen2018neural,sheng2018avatar,kotovenko2019content,hu2020aesthetic}. Intuitively, the category (4) of arbitrary style transfer (AST) is the most advantageous as it is agnostic to $\im{S}$, allowing trained models to be adopted for diverse novel styles without re-training.

Although superior in concept, current AST models are plagued by the issue of imbalanced style transferability (IST), where the stylization intensity of model outputs varies largely across styles $\im{S}$. More importantly, besides the nice results shown in previous works\cite{AdaIn,megenta2,SANet,LinearTransfer}, a large number of stylized images suffer under-stylization (\eg, only the dominant color is transferred) or over-stylization (\ie, content is barely visible) for various $\im{S}$, making them visually undesirable (see samples in Figure~\ref{fig.teaser}). This is validated by our user-study described later in \Cref{subsec.motivation_analysis}, with more than 50\% of stylized images found to be unacceptable, irrespctive of the used AST model. Hence, we are still far from the AST goal --- successfully transferring style from an \emph{arbitrary} image to another. This urges us to systematically study the underlying reasons for \ist and find potential solutions to further boost AST performance in order to generate better stylized images for diverse styles.

In this paper, we make the following contributions. Firstly, we systematically study the \ist problem in AST and discover that the AST style loss is problematic as it fails to reflect human evaluation scores. Secondly, we investigate the AST style loss function, and locate the core reason for \ist to be the way sample-wise style loss is aggregated into a batch loss. Thirdly, we derive the theoretical expectation of a sample-wise style loss as well as its bounds, and use it to propose a new style loss that enables more balanced training across styles. Finally, we conduct extensive AST benchmarking experiments as well as human evaluation to validate the effectiveness of the proposed solution. Results show that \ist issue is indeed greatly mitigated for all tested AST approaches by incorporating the proposed style loss.

The rest of the paper is organized as follows. \Cref{sec.related_work} briefly reviews related AST works. \Cref{sec.motivation} discusses two AST style loss related studies and shows that IST is related to the loss.  \Cref{sec.method} identifies style-agnostic sample weighting in training loss aggregation as the real culprit, derive our new style-aware loss, and validate its effectiveness by repeating the aforementioned two studies. \Cref{sec.evaluation} provides further results of application of the proposed loss to four well-known AST approaches and shows that the \ist issue is largely overcome for all the approaches. Finally, \Cref{sec.conclusion} provides concluding remarks.

\begin{figure*}
    \centering
    \includegraphics[width=0.95\linewidth,trim=0.1cm 0.1cm 0.1cm 0.2cm,clip]{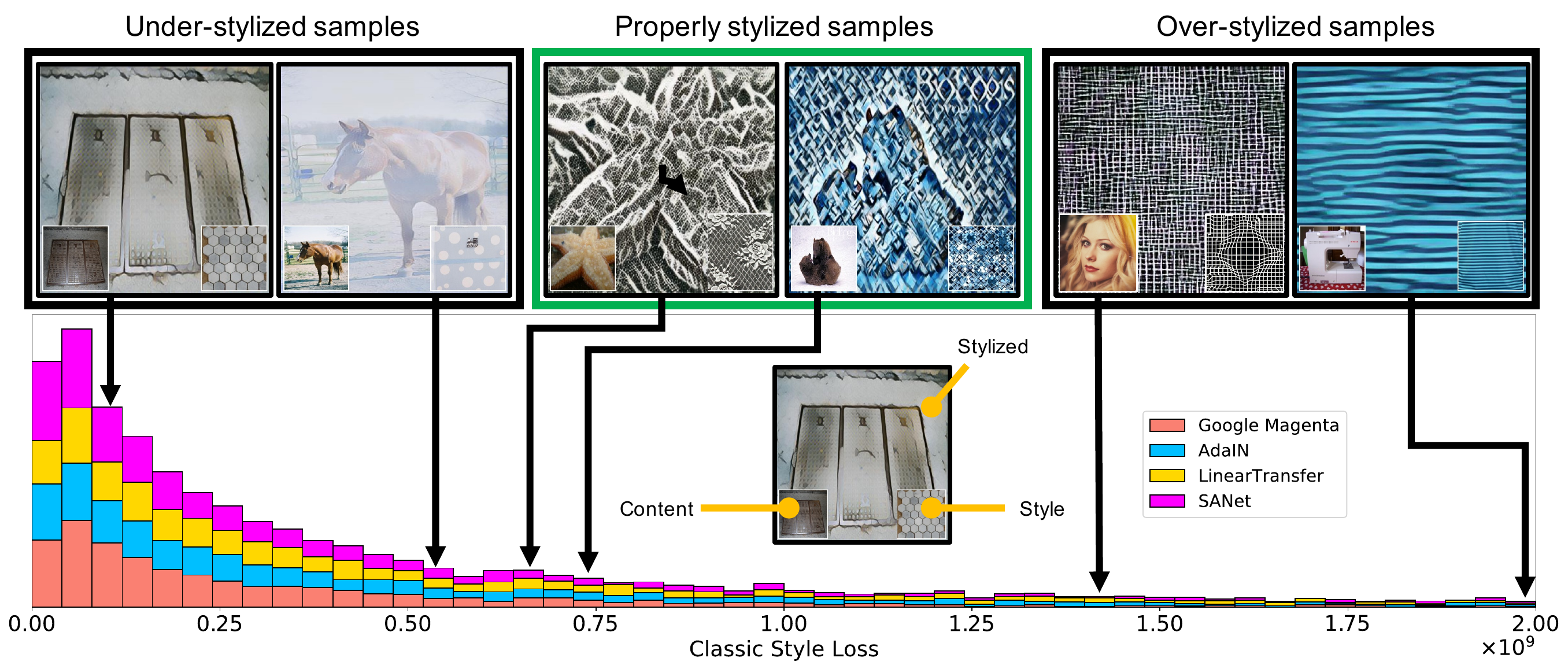}
    \caption{Distribution of classic Gram matrix-based style losses for four AST methods \cite{megenta2,AdaIn,LinearTransfer,SANet}. Smaller loss does not guarantee better style transfer (left two images) while high quality transferred images can have larger style losses (middle two images), with over-stylized images counter-intuitively attaining the highest losses. Zoom in for a better view.}
    \label{fig.loss_histogram1}
    \vspace{-10pt}
\end{figure*}

\section{Related Work}
\label{sec.related_work}

Arbitrary style transfer methods can be classified as either non-parametric~\cite{efros1999texture,efros2001image,kwatra2003graphcut,wei2000fast,WCT,gu2018arbitrary,yoo2019photorealistic,wang2020diversified} or parametric~\cite{SANet,LinearTransfer,AdaIn,megenta2,sheng2018avatar,jing2020dynamic,shen2018neural,kotovenko2019content,wang2020collaborative,svoboda2020two}. Non-parametric methods find similar patches between content and style images, and transfer style based on matched patches. Early methods popularly performed texture synthesis~\cite{efros1999texture,efros2001image,kwatra2003graphcut,wei2000fast}. However, Neural Style Transfer (NST) methods have become mainstream since their inception in~\cite{gatys2016image}. Improvements in the NST framework include multi-level whitening and coloring~\cite{WCT} on VGG~\cite{vgg} features, and feature reshuffling based on patch-based feature similarity~\cite{gu2018arbitrary}.


Parametric AST~\cite{SANet,LinearTransfer,AdaIn,megenta2,sheng2018avatar,kotovenko2019content,shen2018neural,jing2020dynamic,yao2019attention,wang2020collaborative,svoboda2020two} involves optimizing a target function that reflects visual similarity of (1) content between content and stylized images, and (2) style between style and stylized images. This class of AST methods~\cite{SANet,LinearTransfer,AdaIn,megenta2,shen2018neural,jing2020dynamic} typically uses a Gram matrix-based VGG perceptual loss~\cite{gatys2016image,johnson2016perceptual} with a few modifications to the training procedure. AdaIN~\cite{AdaIn} applies an adaptive instance normalization layer on image features. Meta-learning is incorporated for learning style representation in GoogleMagenta~\cite{megenta2} and \cite{shen2018neural}. LinearTransfer~\cite{LinearTransfer} learns a linear transform function from content and style features for stylization. Attention mechanism has also been integrated with content and style feature fusion in SANet~\cite{SANet} and other works~\cite{yao2019attention,svoboda2020two}. Light-weight efficient style transfer has been explored through instance normalization and dynamic convolution~\cite{jing2020dynamic}. Gram matrix-based loss is also widely used in arbitrary video style transfer~\cite{LinearTransfer,chen2017coherent,huang2017real,ruder2018artistic,ruder2016artistic,chen2018stereoscopic}. Methods without Gram matrix-based losses use adversarial~\cite{kotovenko2019content} or reconstruction~\cite{sheng2018avatar} objectives. This paper studies parametric AST methods involving Gram-matrix based losses as listed in \Cref{tab:ast_methods}.

\begin{table}
\centering
\captionsetup{skip=3pt}
\caption{\label{tab:ast_methods}The four studied AST methods with model links.}
\small
\setlength{\tabcolsep}{0.4em} 
\begin{tabular}{l l}
 \toprule
  \bf{AST Method} & \bf{Net Architecture w/ Unique Feature} \\
  \cmidrule(r){1-1} \cmidrule(l){2-2}
         \hyperlink{https://github.com/magenta/magenta/tree/master/magenta/models/arbitrary_image_stylization}{GoogleMagenta}\cite{megenta2}   & ConvNet w/ meta-learned instance norm \\
         \hyperlink{https://github.com/xunhuang1995/AdaIN-style}{AdaIN}\cite{AdaIn}  & Enc.\&Dec. w/ adaptive instance norm \\
         \hyperlink{https://github.com/sunshineatnoon/LinearStyleTransfer}{LinearTransfer}\cite{LinearTransfer} & Enc.\&Dec. w/ linear transform matrix \\
         \hyperlink{https://dypark86.github.io/SANET/}{SANet}\cite{SANet}   & Enc.\&Dec. w/ style attention \\
\bottomrule
\end{tabular}
\vspace{-15pt}
\end{table}

\section{Analysis of the AST Style Loss} \label{sec.motivation}


\subsection{AST Training Loss}

A number of loss functions have been proposed recently for AST training, \eg, discriminator-based adversarial losses \cite{kotovenko2019content} and reconstruction-based loss terms\cite{sheng2018avatar}. However, the original \cite{gatys2016image} NST loss $\loss_{\textrm{NST}}$ is still the one that is the most popularly employed\cite{AdaIn,megenta2,SANet,LinearTransfer,jing2020dynamic,shen2018neural}. As described in \Cref{eq.nst_loss}, it is composed of two terms: $\loss_{\textrm{NST}_c}(\im{C}, \im{P})$ for learning content from $\im{C}$ and $\loss_{\textrm{NST}_s}(\im{S}, \im{P})$ for deriving style from $\im{S}$, with a trade-off factor $\beta$.
\begin{equation}\label{eq.nst_loss}
    \loss_{\textrm{NST}} = \loss_{\textrm{NST}_c}(\im{C}, \im{P}) + \beta \loss_{\textrm{NST}_s}(\im{S}, \im{P})
\end{equation}
One typically needs an ImageNet\cite{deng2009imagenet} pretrained VGG\cite{vgg} network $\feat$ for extracting features from $\im{C}$, $\im{S}$, and $\im{P}$. Next, the content loss is calculated by comparing the features of $\im{P}$ and $\im{C}$, and the style term is calculated by comparing the Gram matrices $\gram$ of the features of $\im{P}$ and $\im{S}$, as $\gram$ is known\cite{gatys2016image,li2017demystifying} to be effective in deriving style information. In practice, the style and content terms are calculated for features from several layers 
and aggregated using a weighted sum (weights $w^l$ are typically set as ones) across layers. The following equations summarize the loss calculation, where MSE is the mean squared error.
\begin{align}
    \loss_{\textrm{NST}_c}^l(\im{C}, \im{P}) &= \mse(\feat^l(\im{C}), \feat^l({\im{P}})) \label{eq.layerwise_content_loss} \\
    \loss_{\textrm{NST}_s}^l(\im{S}, \im{P}) &= \mse(\gram\circ\feat^l(\im{S}), \gram\circ\feat^l({\im{P}})) \label{eq.layerwise_style_loss} \\
    \loss_{\textrm{NST}_c}(\im{C}, \im{P}) &= \textstyle\sum_{l\in\mathbb{L}_{\textrm{NST}_c}}w_c^l\cdot\loss_{\textrm{NST}_c}^l(\im{C},\im{P}) \label{eq.classic_content_loss} \\
    \loss_{\textrm{NST}_s}(\im{S}, \im{P}) &= \textstyle\sum_{l\in\mathbb{L}_{{\textrm{NST}}_s}}w_s^l\cdot\loss_{\textrm{NST}_s}^l(\im{S}, \im{P}) \label{eq.classic_style_loss}
\end{align}

\subsection{Analysis}
\label{subsec.motivation_analysis}

The \ist issue could be intuitively attributed to the ``naturally higher'' difficulty of transferring certain styles compared to others. In order to study \ist systematically, we calculate content and style losses for 20,000 randomly sampled ImageNet\cite{deng2009imagenet} images stylized with images from the Describable Textures Dataset (DTD)\cite{dtd} and pretrained models listed in \Cref{tab:ast_methods}. 
Ideally, under- and over-stylized $\im{P}$ are those with low content and low style losses, respectively. On analyzing the samples, however, we find that the former relationship is valid but the latter is not --- over-stylized $\im{P}$ typically attain (sometimes significantly) higher style losses than understylized samples. In the following studies, we examine the distribution of the style loss and its correlation with visual perception of stylization quality.



\vspace{5pt}\noindent\textbf{Study I: AST Style Loss Distribution.} We compute the empirical distribution of style losses for the models listed in \Cref{tab:ast_methods} and inspect stylized samples belonging to different sections of the distribution -- low, moderate, and high style losses. We use a VGG-16 model pretrained on ImageNet as the feature extractor, and calculate the style loss (see \Cref{eq.classic_style_loss}) using layers $\feat^l$ in the conventional style layer set~\cite{johnson2016perceptual} $\mathbb{L}_{{\textrm{AST}}_s} = \{\feat_{b1}^{r2}, \feat_{b2}^{r2}, \feat_{b3}^{r3}, \feat_{b4}^{r4}\}$, where $\feat_{bi}^{rj}$ denotes the $j$-th ReLU layer in $i$-th convolutional block of VGG-16. The AST style loss is thus restated as below.
\begin{align}
\loss_{\textrm{AST}_s}^l(\im{S}, \im{P}) &= \mse(\gram\circ\feat^l(\im{S}), \gram\circ\feat^l({\im{P}})) \label{eq.layerwise_ast_style_loss} \\
\loss_{\textrm{AST}_s}(\im{S}, \im{P}) &=\textstyle \sum_{l\in\mathbb{L}_{{\textrm{AST}}_s}}w_s^l\cdot\loss_{\textrm{AST}_s}^l(\im{S}, \im{P}) \label{eq.classic_ast_style_loss}
\end{align}
\Cref{fig.loss_histogram1} summarizes our findings. Despite large differences among the tested methods, (1) their $\loss_{\textrm{AST}_s}$ distributions are similar, and (2) $\loss_{\textrm{AST}_s}$ does not reflect stylization quality: under-stylized samples attain lower loss values than over-stylized ones. Similar conclusions can be drawn for VGG-19-based $\loss_{\textrm{AST}_s}$, as shown in the supplementary material.

\vspace{5pt}\noindent\textbf{Study II: Classic AST Style Loss versus Human Score.} Study I has revealed a counter-intuitive lack of relationship between style transfer quality and conventional AST style loss. In this study, we further investigate this issue by conducting a human study to assess the correlation between $\loss_{\textrm{AST}_s}$ and human perception. Specifically, we requested five volunteers to manually annotate AST samples by partitioning the samples produced in Study I into five random disjoint subsets. Each sample was presented as a tuple of $(\im{S}, \im{P})$ and had to be annotated as ``Good'' (-1), ``OK'' (0), or ``Bad'' (1), in decreasing order of stylization quality. The annotators were not given additional instructions and were told to classify the samples based on their own perception.

\begin{figure}[!htbp]
    \centering
    \includegraphics[width=0.9\linewidth,trim=0.05cm 1.75cm 0.75cm 0.2cm,clip]{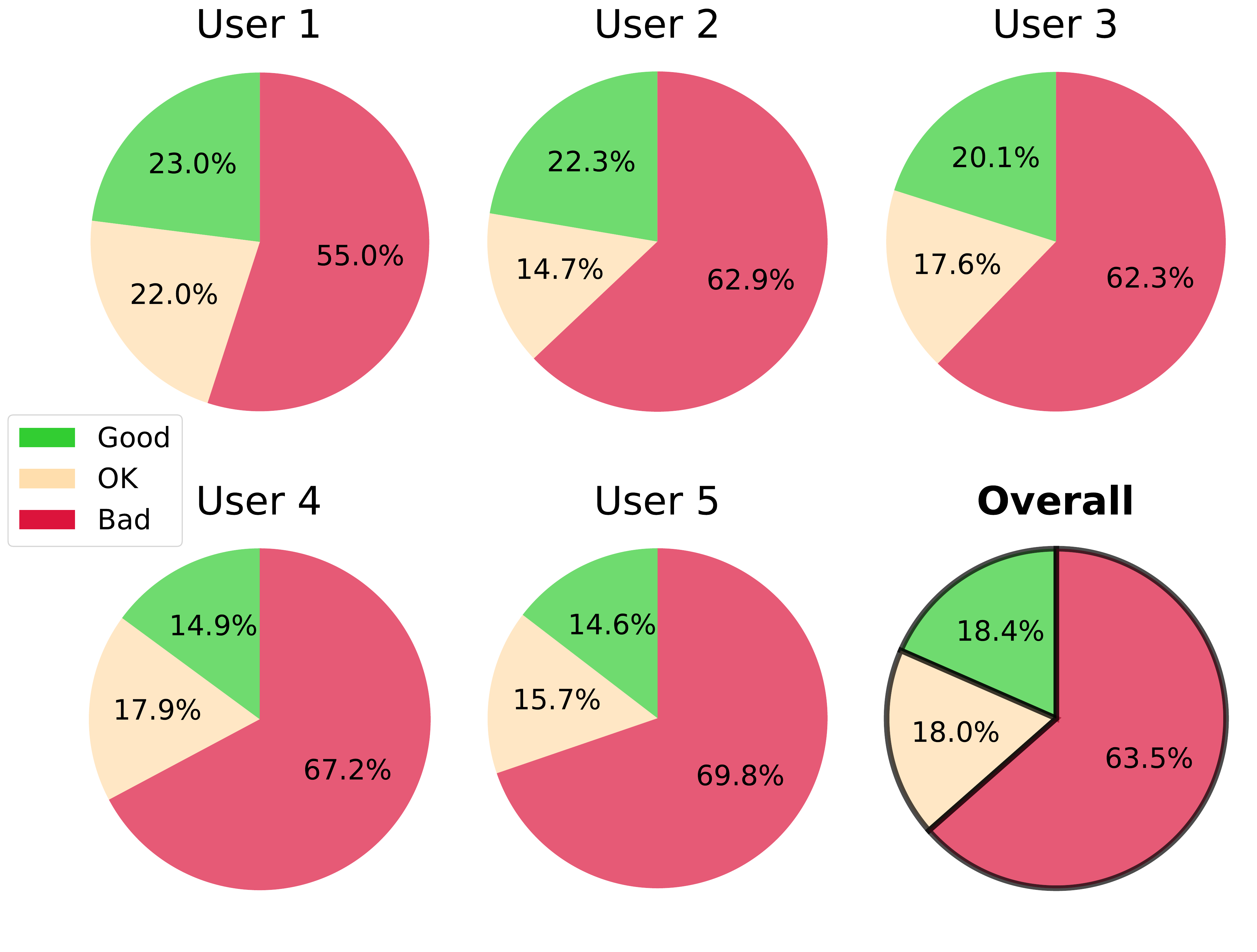}
    \caption{Statistics of human perception of stylization quality as assessed in Study II.}
    \label{fig.human_eval}
    \vspace{-10pt}
\end{figure}

\Cref{fig.human_eval} shows the statistics of the collected annotations. We compute Pearson correlation between the human scores and corresponding style losses. \Cref{tab.classic_loss_human_coor} summarizes the results, showing that the conventional style loss not only fails to reflect human perception but is also negatively correlated. Hence, $\loss_{\textrm{AST}_s}$ (as defined in \Cref{eq.classic_ast_style_loss}) is inappropriate for the AST task. Furthermore, the negative correlation indicates that this style loss penalizes over-stylized samples more than under-stylized ones --- contrary to what one would expect for a good AST style loss.

\begin{table}[!htbp]
    \centering
    \small
    \caption{Pearson correlation between the classic AST style loss ($\loss_{\textrm{AST}_s}$) and human score ($h$). $\feat_{bi}^{rj}$ indicates the the $j$-th ReLU layer in $i$-th convolutional block of VGG-16.}
    \setlength{\tabcolsep}{0.75em} 
    \begin{tabular}{cccccc}
        \toprule
         & \multicolumn{5}{c}{\textbf{Correlation Score} $\rho(h, \loss_s)$} \\
         \cmidrule(l){2-6}
        \textbf{Volunteer} & $\loss_s^{\feat_{b1}^{r2}}$ & $\loss_s^{\feat_{b2}^{r2}}$ & $\loss_s^{\feat_{b3}^{r3}}$ & $\loss_s^{\feat_{b4}^{r4}}$ & $\loss_{\textrm{AST}_s}$ \\
        \cmidrule(lr){1-1} \cmidrule(lr){2-2} \cmidrule(lr){3-3} \cmidrule(lr){4-4} \cmidrule(lr){5-5} \cmidrule(lr){6-6}
        1 & -0.166 & -0.153 & -0.186 & -0.114 & -0.167  \\
        2 & -0.163 & -0.141 & -0.139 & -0.074 & -0.150  \\
        3 & -0.184 & -0.161 & -0.147 & -0.177 & -0.163  \\
        4 & -0.194 & -0.147 & -0.127 & 0.006 & -0.138  \\
        5 & -0.214 & -0.224 & -0.192 & -0.070 & -0.219  \\
        \cmidrule(lr){1-6}
        Average & -0.184 & -0.165 & -0.158 & -0.088 & -0.167\\
        \bottomrule
    \end{tabular}
    \label{tab.classic_loss_human_coor}
    \vspace{-10pt}
\end{table}




\section{A New Blalanced AST Style Loss}
\label{sec.method}

The conventional AST style loss defined in \Cref{eq.classic_ast_style_loss} reuses the classic NST style loss, which has been shown to work in practice in several previous works \cite{AdaIn,megenta2,SANet,LinearTransfer,shen2018neural}. However, results of studies I and II reveal that this AST style loss is problematic: trained models work partially but suffer from Imbalanced Style Transferability (\ist), \ie, under- or over-stylization for various styles with loss values that do not reflect stylization quality. In this section, we first identify the core issue in the AST style loss by viewing AST from a multi-task learning point-of-view. We then propose a simple yet effective solution to mitigate the problem. 



\subsection{Identifying the Real Problem with AST Loss}

The conventional AST style loss (also the classic NST style loss) used in studies I and II is a sample-wise loss. However, in order to ascertain the cause of the aforementioned issue, it is important to inspect how it is used in training --- it needs to be aggregated into a batch-wise loss as
\begin{equation}\label{eq.ast_batch_loss}
\loss_{\textrm{AST}_s} ^{\textrm{Batch}} = \sum_{k\in\{1,\cdots, B\}}{1\over B}\cdot\loss_{\textrm{AST}_s}(\im{S_k}, \im{P_k}) 
\end{equation}
where $B$ is the batch size. While it is typical to average sample-wise losses into batch-losses in this fashion, this protocol is not suitable for AST. This is because the AST learning setup resembles multi-task learning, where each batch has $B$ tasks -- one for each input style. The overall multi-task loss can be written as
\begin{equation}\label{eq.ast_batch_loss2}
\loss_{\textrm{AST}_s}^{\textrm{Multitask}} = \sum_{k\in\{1,\cdots, B\}}\lambda_k\cdot\loss_{\textrm{AST}_s}(\im{S_k}, \im{P_k}) 
\end{equation}
where $\lambda_k$ is typically a task-specific contribution factor. Comparing Equations \eqref{eq.ast_batch_loss} and \eqref{eq.ast_batch_loss2}, it is clear that the AST style loss is a special case of the multi-task loss when $\lambda_k=1/B$ for each $k$. However, this equal-task-weight setting is known to be problematic in multi-task learning unless all task losses are within similar dynamic ranges\cite{chen2018gradnorm,kendall2018multi}.

In case of AST, style losses for different style images can differ by more than 1,000 times for both randomly initialized and fully trained AST models. Consequently, styles with small or large dynamic loss ranges are under- or over-stylized, respectively. Although $\lambda_k=1/B$ works for some style images, generating nice stylization results for them, this setting is unsuitable for the general AST problem and is the root cause of the discrepancy between stylization quality and loss values. Therefore, we should neither simply aggregate the sample-wise losses to form a batch-wise loss nor directly compare losses from different styles.

\subsection{A New Balanced AST Style Loss}
\label{subsec:new_loss}

The multi-task view discussed in the previous section implies that the \ist problem could be resolved by assigning each style transfer task in a batch the ``right'' task weight. Hence, we seek to formulate a balanced AST style loss as
\begin{equation}\label{eq.layerwise_ast_style_loss_new}
   \hat{\loss}_{\textrm{AST}_{s}}^l(\im{S}, \im{P}) = {\loss_{\textrm{AST}_{s}}^l(\im{S}, \im{P})\over {\range^l(\im{S}, \im{P})} }
\end{equation}
where $\range^l(\im{S}, \im{P})$ is the appropriate task-dependent normalization term that needs to be determined (where $\lambda_k = 1 / \range^l(\im{S}, \im{P})$). An intuitive approach to achieve this is to adopt automatic task loss-weight tuning methods from the multi-task literature~\cite{kendall2018multi,guo2018dynamic,chen2018gradnorm,liu2019end}. However, these methods require estimation of statistics (\eg, gradient norms) for all tasks in multiple iterations (if not continuously), which is infeasible for AST as the tasks change across batches and the number of $(\im{S}, \im{P})$ combinations is potentially infinite. Therefore, weight tuning approaches are not suitable for AST. Furthermore, the choice of $\range^l(\im{S}, \im{P})$ is limited to something that could be computed without historical data.

We start with deriving the theoretical upper- and lower-bounds for the classic AST layerwise style loss (\Cref{eq.layerwise_ast_style_loss}) as shown in Equations \eqref{eq.upperbound} and \eqref{eq.lowerbound}, respectively:
\vspace{-5pt}
\begin{align}
   \!\! \sup\{\loss_{\textrm{AST}_s}^l(S,P)\} \!\!&=\!\! {\|\gram\circ\feat^l(S)\|^2 + \|\gram\circ\feat^l(P)\|^2 \over N^l} \label{eq.upperbound} \\
   \!\! \inf\{\loss_{\textrm{AST}_s}^l(S,P)\} \!\!&=\! \!{(\|\gram\circ\feat^l(S)\| - \|\gram\circ\feat^l(P)\|)^2 \over N^l} \label{eq.lowerbound}
\end{align}
where $N^l$ is a constant that is equal to the product of spatial dimensions of the feature tensor at layer $l$. The detailed derivations can be found in the supplementary material.

In order to mitigate Imbalanced Style Transferability, we propose a new style-balanced loss $\hat{\loss}_{\textrm{AST}_{s}}^l$ by normalizing the style loss of each AST task with its supremum as:
\begin{equation}\label{eq.layerwise_ast_style_loss_new2}
    \hat{\loss}_{\textrm{AST}_{s}}^l(\im{S}, \im{P}) = {\loss_{\textrm{AST}_{s}}^l(\im{S}, \im{P})\over {\sup\{\loss_{\textrm{AST}_s}^l(S,P)\} } }
\end{equation}

\subsection{Analysis and Validation of Effectiveness}

We conduct three studies to analyze and validate the correctness and effectiveness of our new loss (\Cref{eq.layerwise_ast_style_loss_new2}).

\vspace{5pt}\noindent\textbf{Study III: Relationship between} $\loss_{\textrm{AST}_{s}}^l(\im{S}, \im{P})$ \textbf{and} $ {\sup\{\loss_{\textrm{AST}_s}^l(S,P)\} }$\textbf{.}\quad It is important to establish this relationship to ensure that ${\sup\{\loss_{\textrm{AST}_s}^l(S,P)\} }$ is a suitable normalization term. Specifically, the relationship has to be close to linear to balance all the training tasks by ensuring that all the training tasks have the same upper bound of $1$.

\begin{figure}
    \centering

    \includegraphics[width=\linewidth,trim=0.15cm 0.1cm 0.25cm 0.1cm,clip]{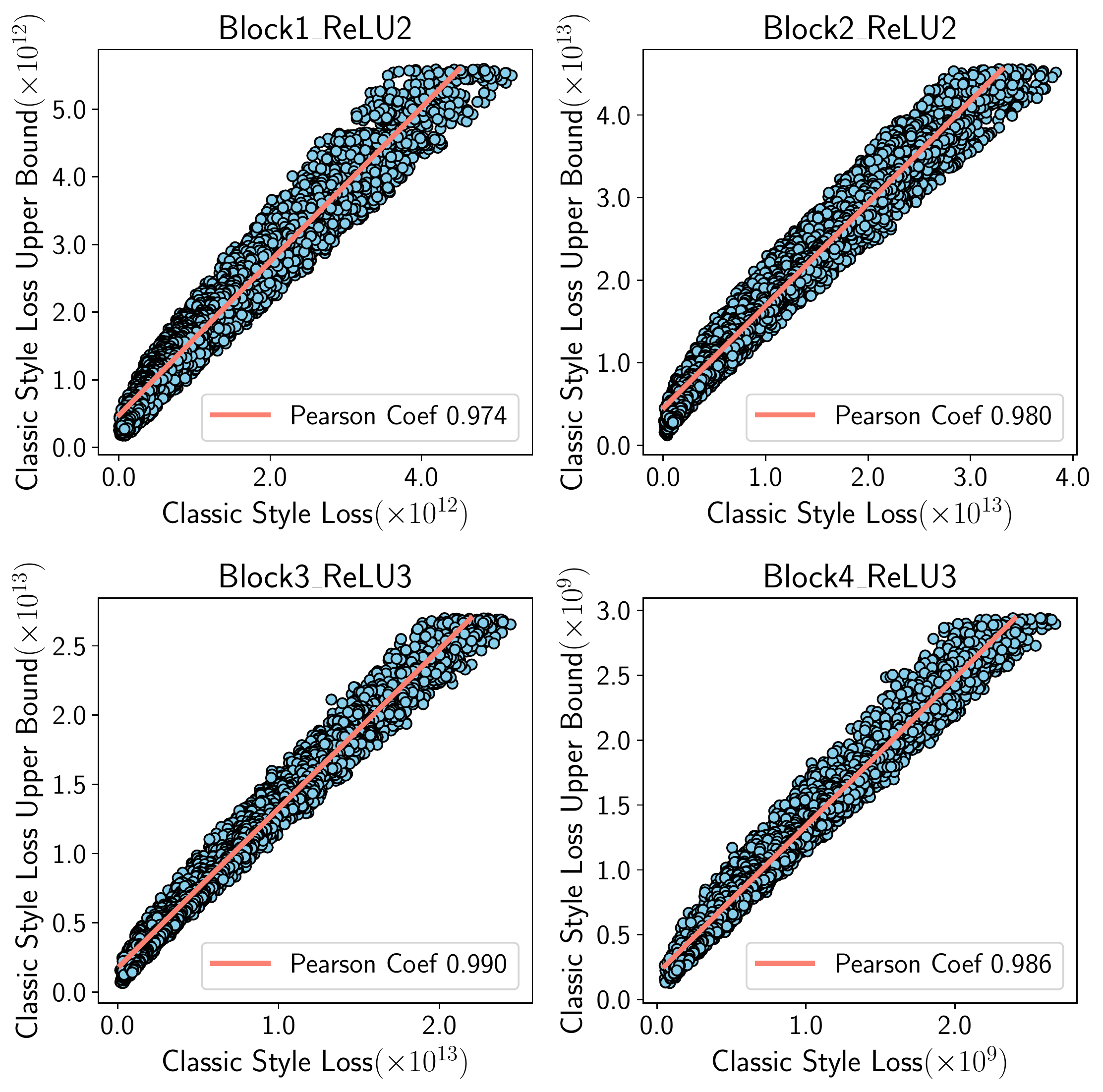}

    \caption{Relationship between the classic AST style loss $\loss_{\textrm{AST}_{s}}^l(\im{S}, \im{P})$ and $ {\sup\{\loss_{\textrm{AST}_s}^l(S,P)\} }$. The subplots correspond to the four VGG-16 style layers used in analysis.}
    \label{fig.corr_before_norm}
    \vspace{-10pt}
\end{figure}

We randomly sample 200,000 pairs of images from the Painter by Numbers (PBN) dataset~\cite{pbn}. For each pair, we compute the classic layerwise style loss $\loss_{\textrm{AST}_{s}}^l(\im{S}, \im{P})$ using the VGG-16 style layer set~\cite{johnson2016perceptual} (\Cref{eq.classic_ast_style_loss}), and its upper-bound (\Cref{eq.upperbound}). \Cref{fig.corr_before_norm} provides a scatter plot of the two terms, where each dot is a sample and the red line is the linear fit of all samples, showing that $\loss_{\textrm{AST}_{s}}^l(\im{S}, \im{P})$ and $ {\sup\{\loss_{\textrm{AST}_s}^l(S,P)\} }$ are strongly correlated.



\begin{figure*}[t]
    \centering
    \includegraphics[width=0.95\linewidth,trim=0.1cm 0.1cm 0.1cm 0.2cm,clip]{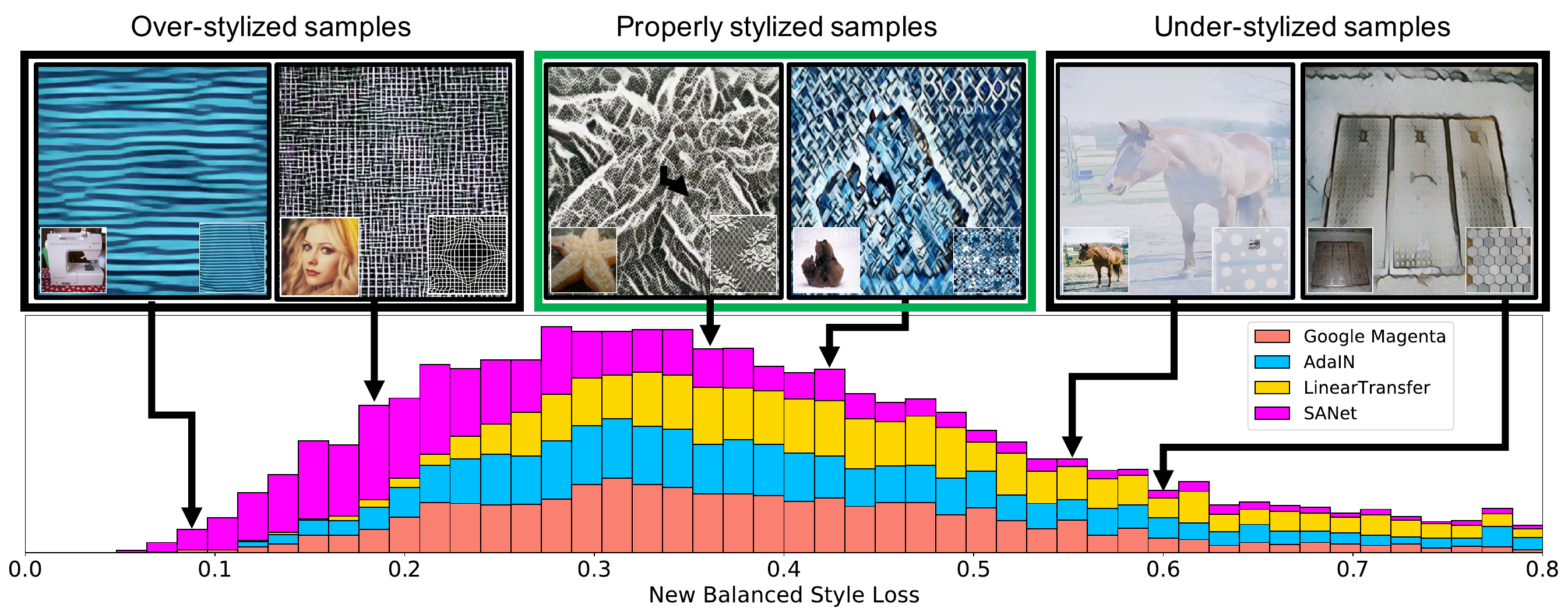}
    \caption{Distribution of our style-balanced loss for four AST methods \cite{megenta2,AdaIn,LinearTransfer,SANet}. Small loss values indicate over-stylization while large values correspond to under-stylization, with properly-stylized samples in the middle, as expected.}
    \label{fig.loss_histogram2}
    \vspace{-10pt}
\end{figure*}

\vspace{5pt}\noindent\textbf{Study IV: Distribution of the New Balanced AST Style Loss.}\quad As previously noted in Study I, under-stylized samples typically attain lower loss values than over-stylized ones in the classic AST style loss. Here we verify whether our new AST style loss fixes this issue. We reuse the data from Study I for the four tested AST approaches (see \Cref{tab:ast_methods}), and compute the corresponding new style loss distributions as shown in \Cref{fig.loss_histogram2}. Results show that over-stylized samples now attain lower loss values than under-stylized ones under the new AST style loss $\hat{\loss}_{\textrm{AST}_{s}}$. 

\vspace{5pt}\noindent\textbf{Study V: New AST Style Loss vs. Human Score.}\quad  We investigate the relationship between our new balanced AST style loss $\hat{\loss}_{\textrm{AST}_{s}}$ and human scores for the samples generated in Study II. These results are presented in \Cref{tab.normed_loss_human_coor}. Unlike the negative correlation between the classic AST style loss and human scores (see \Cref{tab.classic_loss_human_coor}), the new balanced AST style loss is positively correlated with human scores. Despite the treatment of ``OK'' annotations as zero scores in human study analysis, this is a strong indication that the proposed new AST style loss aligns much better with human scores than the classic AST style loss. 

\begin{table}
    \centering
    \small
    \caption{Pearson correlation between the new AST style loss ($\hat{\loss}_{\textrm{AST}_s}$) and human scores ($h$). $\feat_{bi}^{rj}$ indicates the the $j$-th ReLU layer in $i$-th convolutional block of VGG-16.}\label{tab.normed_loss_human_coor}
    \setlength{\tabcolsep}{0.75em} 
    \begin{tabular}{cccccc}
        \toprule
         & \multicolumn{5}{c}{\textbf{Correlation Score} $\rho(h, \hat{\loss}_s)$} \\
         \cmidrule(l){2-6}
        \textbf{Volunteer} & $\hat{\loss}_s^{\feat_{b1}^{r2}}$ & $\hat{\loss}_s^{\feat_{b2}^{r2}}$ & $\hat{\loss}_s^{\feat_{b3}^{r3}}$ & $\hat{\loss}_s^{\feat_{b4}^{r3}}$ & $\hat{\loss}_{\textrm{AST}_s}$ \\ \cmidrule(lr){1-1} \cmidrule(lr){2-2} \cmidrule(lr){3-3} \cmidrule(lr){4-4} \cmidrule(lr){5-5} \cmidrule(lr){6-6}
        1 & -0.086 & 0.095 & 0.188 & 0.121 & 0.152  \\
        2 & 0.002 & 0.001 & 0.098 & 0.122 & 0.124  \\
        3 & -0.087 & 0.106 & 0.217 & 0.217 & 0.228  \\
        4 & -0.067 & -0.089 & 0.053 & 0.210 & 0.149  \\
        5 & -0.141 & -0.118 & 0.112 & 0.208 & 0.171  \\
        \cmidrule(lr){1-6}
        Average & -0.076 & -0.001 & 0.134 & 0.176 & 0.165\\
        \bottomrule
    \end{tabular}
    \vspace{-10pt}
\end{table}

\section{Experimental Evaluation}
\label{sec.evaluation}
In this section, we provide qualitative and quantitative evaluation of the proposed new loss and its benefits in comparison with the conventional AST style loss.


\begin{table}
    \centering
    \small
    \caption{The AST models used in experimental evaluation. ``3P'' indicates publicly available pretrained models, while ``1P'' shows the settings used in our experiments. We train all models using VGG-16-based losses for consistency.\vspace{-3pt}}
    \label{tab:eval_models}
    \setlength{\tabcolsep}{0.3em} 
    \begin{tabular}{lccc}
         \toprule
         \bf{Model Name} & \bf{AST Method} & \bf{AST Style Loss} & \bf{1P/3P} \\
         \cmidrule(r){1-1} \cmidrule(lr){2-2} \cmidrule(lr){3-3} \cmidrule(l){4-4}
         \hyperlink{https://github.com/magenta/magenta/tree/master/magenta/models/arbitrary_image_stylization}{GoogleMagenta} &  GoogleMagenta & Classic + VGG-16 & 3P \\
         OurGM &  GoogleMagenta & Classic + VGG-16 & 1P \\
         OurBalGM & GoogleMagenta & Balanced + VGG-16 & 1P \\
         \cmidrule(r){1-1} \cmidrule(lr){2-2} \cmidrule(lr){3-3} \cmidrule(l){4-4}
         \hyperlink{https://github.com/xunhuang1995/AdaIN-style}{AdaIN} &  AdaIN & Classic + VGG-19 & 3P \\
         OurAI &  AdaIN & Classic + VGG-16 & 1P \\
         OurBalAI & AdaIN & Balanced + VGG-16 & 1P \\
         \cmidrule(r){1-1} \cmidrule(lr){2-2} \cmidrule(lr){3-3} \cmidrule(l){4-4}
         \hyperlink{https://github.com/sunshineatnoon/LinearStyleTransfer}{LinearTransfer} &  LinearTransfer & Classic + VGG-19 & 3P \\
         OurLT &  LinearTransfer & Classic + VGG-16 & 1P \\
         OurBalLT &  LinearTransfer & Balanced + VGG-16 & 1P \\
         \cmidrule(r){1-1} \cmidrule(lr){2-2} \cmidrule(lr){3-3} \cmidrule(l){4-4}
         \hyperlink{https://dypark86.github.io/SANET/}{SANet} & SANet & Classic + VGG-19 & 3P \\
         OurSAN&  SANet & Blassic + VGG-16 & 1P \\
         OurBalSAN &  SANet & Balanced + VGG-16 & 1P \\
         \bottomrule
    \end{tabular}
    \vspace{-10pt}
\end{table}

\subsection{Experiment Settings}

We reuse the pretrained models listed in \Cref{tab:ast_methods} and train additional models with either the classic or the new style loss in order to validate the generalizability of the improvements due to the latter. These models are listed in \Cref{tab:eval_models}. We use content images from MS-COCO~\cite{mscoco} and style images from Painter by Numbers~\cite{pbn} to train the models. The same four layers ($\mathbb{L}_{{\textrm{AST}}_s} = \{\feat_{b1}^{r2}, \feat_{b2}^{r2}, \feat_{b3}^{r3}, \feat_{b4}^{r4}\}$) of the ImageNet-trained VGG-16 model that were used in Studies I--V were employed here to calculate style losses for training. The $\feat_{b3}^{r3}$ layer was reused to calculate content losses, following previous works~\cite{gatys2016image,johnson2016perceptual}. In order to conduct fair comparison of models, we pick style vs. content trade-off weights $\beta$ in the overall loss (\Cref{eq.nst_loss}) that ensure that the magnitudes of the style and content losses are similar. We use the same optimizers that were used by the authors of the four models as reported in literature. We use ``1P'' models for the classic AST style loss related comparisons. We do not backpropagate gradient through the denominator to obtain more stable training and better results.


\subsection{Qualitative Evaluation}

\begin{figure*}
    \centering
    \includegraphics[width=\linewidth,trim=0.95cm 0.3cm 0 0.25cm,clip]{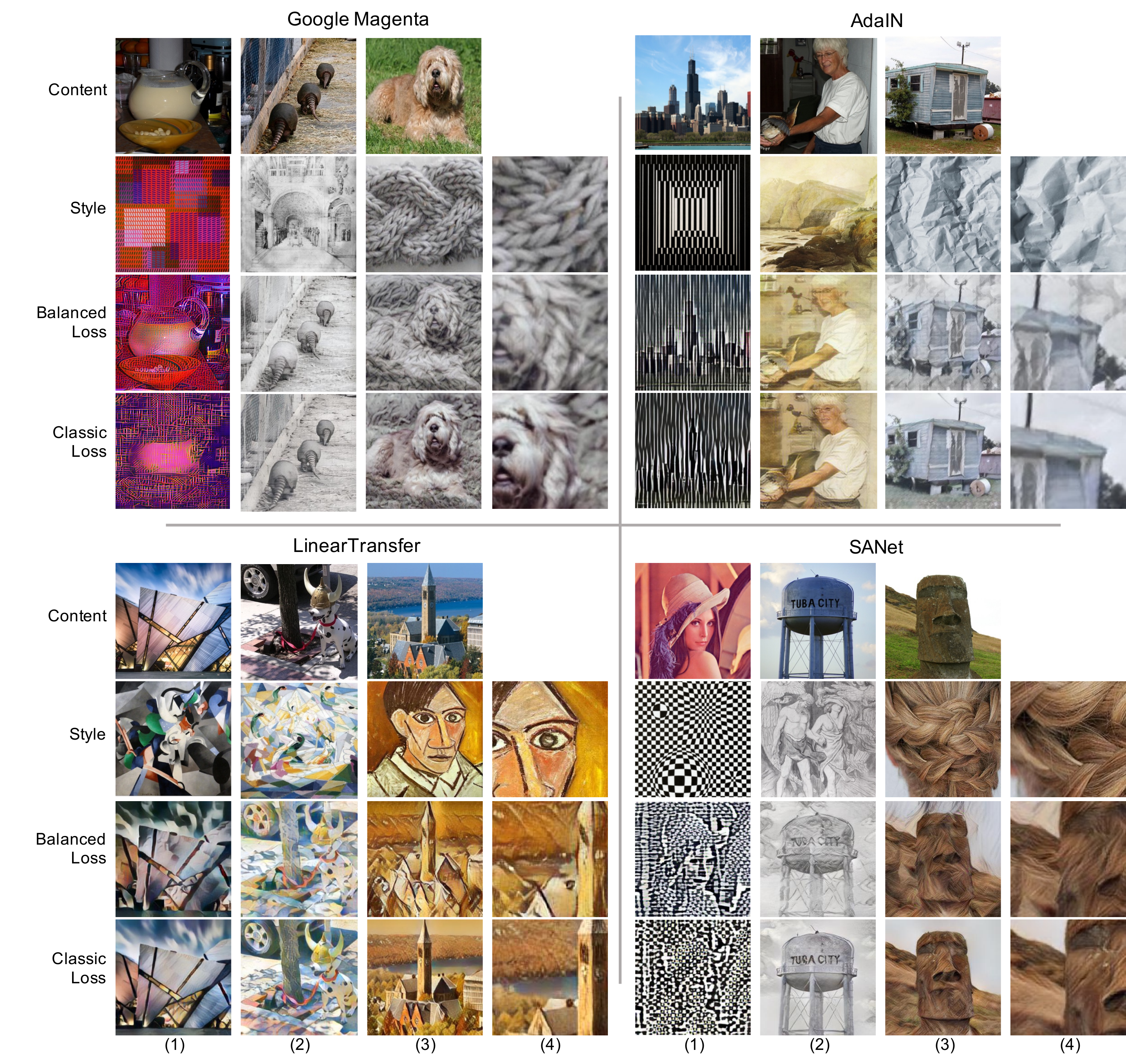}
    \caption{Improvements in cases that were under- or over-stylized due to the classic loss. In the former case, local and global texture of stylized images are closer to style images when our new loss is used, e.g., columns (3) and (4) for Google Magenta. In the latter case, original content is more visible in stylized images when our loss is used, e.g., column (1) of AdaIN.}
    \label{fig.qualitative_of_improving}
    \vspace{-10pt}
\end{figure*}

\noindent\textbf{Comparison with Classic Loss.}\quad \Cref{fig.qualitative_of_improving} shows a few visual examples of under- and over-stylization that are mitigated by training models with our new balanced style loss. As evident, our loss is effective in both cases and generalizes across all tested AST models, providing a better trade-off between content and style. Results of under-stylized samples show that our loss helps in capturing both global and low-level texture-related style information where models trained with the classic loss only contain style-color. Results also show that our loss can preserve more content in cases of over-stylization. While content is completely unrecognizable in over-stylized images due to the classic loss, our loss produces both visible content and proper stylization. More results are presented in  supplementary material.

\begin{figure}
    \centering
    \includegraphics[width=\linewidth,trim=0.5cm 0.45cm 0 0.45cm,clip]{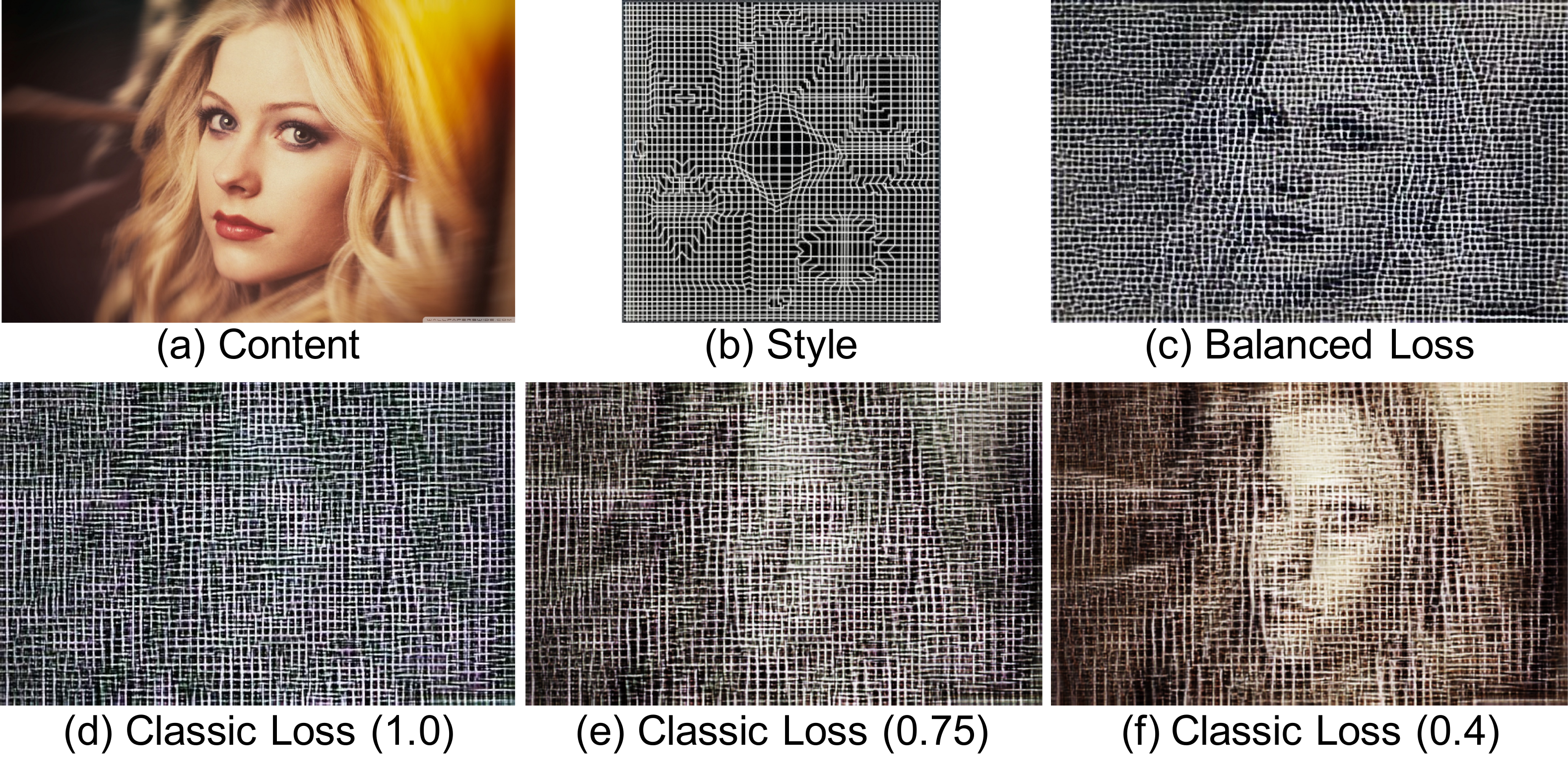}
    \caption{Our new balanced style loss is more effective in mitigating the over-stylization issue than the classic style interpolation approach using the GoogleMagenta solution. The style interpolation coefficient is presented in braces.  }
    \label{fig.comparison_with_interpolation}
    \vspace{-5pt}
\end{figure}

\noindent\textbf{Comparison with Style Interpolation.}\quad Style interpolation is a common method for fusing styles by combining style features of different images before decoding\cite{AdaIn,megenta2,LinearTransfer,shen2018neural}. Typically, an interpolation coefficient is employed to control the contributions of different styles. Since over-stylization is mainly about transferring too much style, one plausible remedy is to apply style interpolation between the style and content (\ie treating the content image as a new style).
However, as shown in \Cref{fig.comparison_with_interpolation}, it is not very effective. In contrast, stylization using our balanced loss (\Cref{fig.comparison_with_interpolation}(c)) provides superior results in better preserving content while properly transferring style. Last but not least, finding a ``good'' interpolation coefficient is not a trivial task, since it will be different for different styles. 

\subsection{Quantitative Evaluation}

\vspace{5pt}\noindent\textbf{AST Style Loss Comparison.}\quad We compute the classic and the new balanced AST style losses for all the testing samples for each AST model. Results are presented in \Cref{tab:loss_comparison}. As evident, our models trained with the new balanced loss always attain significantly lower overall loss values. This is because the new loss allows more tasks ($(\im{S}, \im{C})$ combinations) to be trained fairly, achieving lower classic style losses for the same pairs of content and style images.



\begin{table}
    \centering
    \caption{Loss comparison between models trained with the classic loss and those trained with our new loss. Models trained with our loss always attain lower overall loss values.\vspace{-3pt}}
    \label{tab:loss_comparison}
    \small
    \setlength{\tabcolsep}{0.9em}
    \begin{tabular}{lcc}
    \toprule
    \bf{Model Name} & \bf{Classic Loss} & \bf{New Balanced Loss} \\
    \cmidrule(r){1-1} \cmidrule(lr){2-2} \cmidrule(l){3-3}
         GoogleMagenta &  4.69 $\times 10^8$ & 0.36 \\
         OurGM & 3.40 $\times 10^8$ & 0.33 \\ 
         OurBalGM & \textbf{3.35 $\mathbf{\times 10^8}$} & \textbf{0.22} \\
         \cmidrule(r){1-1} \cmidrule(lr){2-2} \cmidrule(l){3-3}
         AdaIN & 7.05 $\times 10^8$ & 0.33\\
         OurAI & 6.62 $\times 10^8$ & 0.43 \\
         OurBalAI & \textbf{6.58 $\mathbf{\times 10^8}$} & \textbf{0.31} \\
         \cmidrule(r){1-1} \cmidrule(lr){2-2} \cmidrule(l){3-3}
         LinearTransfer & 6.11 $\times 10^8$ & 0.33 \\
         OurLT & 6.78 $\times 10^8$ & 0.47 \\
         OurBalLT & \textbf{4.27 $\mathbf{\times 10^8}$} & \textbf{0.25} \\
         \cmidrule(r){1-1} \cmidrule(lr){2-2} \cmidrule(l){3-3}
         SANet & 5.00 $\times 10^8$ & 0.28 \\
         OurSAN & 5.25 $\times 10^8$ & 0.41 \\
         OurBalSAN & \textbf{4.03 $\mathbf{\times 10^8}$} & \textbf{0.21} \\ 
    \bottomrule
    \end{tabular}
\end{table}

\vspace{5pt}\noindent\textbf{Deception Rate.}\quad Sanakoyeu \textit{et al.}\cite{sanakoyeu2018style} introduced Deception Rate as a metric for evaluating style transfer quality. It is defined as the success rate of stylized images at deceiving an artist classification model, such that the same artist is predicted for both the style image and the stylized image.

We generate 5,000 stylized images for each AST model with content images from the ImageNet test set and style images from the Painter by Numbers (PBN) test set in order to ensure that the images and styles used in this experiment do not overlap with the training dataset. Furthermore, we exclude style images from artists who were seen in the training data or have less than 30 paintings in the test set. This results in 1,798 style images (paintings) from 34 artists. Content-style pairs were randomly sampled to generate the stylized images for evaluation.

\begin{table}
    \centering
    \caption{Deception Rate (\%) for models trained with classic and our balanced style losses, shown in columns ``Classic Loss'' and ``Bal. Loss'', respectively. ``Imp.'' and ``RImp.'' show absolute and relative improvements, respectively.\vspace{-3pt}}
    \label{tab.deception_rate}
    \small
    \setlength{\tabcolsep}{0.45em}
    \begin{tabular}{lcccc}
    \toprule
    \bf{AST Method} & \bf{Classic Loss} & \bf{Bal. Loss} & \bf{Imp.} & \bf{RImp.} \\
    \cmidrule(r){1-1} \cmidrule(lr){2-2} \cmidrule(lr){3-3} \cmidrule(lr){4-4} \cmidrule(l){5-5}
         GoogleMagenta &  16.84\% & 35.38\% & 18.54\%  & 110\% \\
         AdaIN & 10.78\% & 22.18\% & 11.40\%  & 106\% \\
         LinearTransfer & 18.58\% & 39.18\% & 20.60\%  & 111\% \\
         SANet & 18.08\% & 33.16\% & 15.08\%  & ~~83\% \\
    \bottomrule
    \end{tabular}
    \vspace{-10pt}
\end{table}

\begin{figure}
    \centering
    \includegraphics[width=0.9\linewidth,trim=0.5cm 2.5cm 0.2cm 0.5cm,clip]{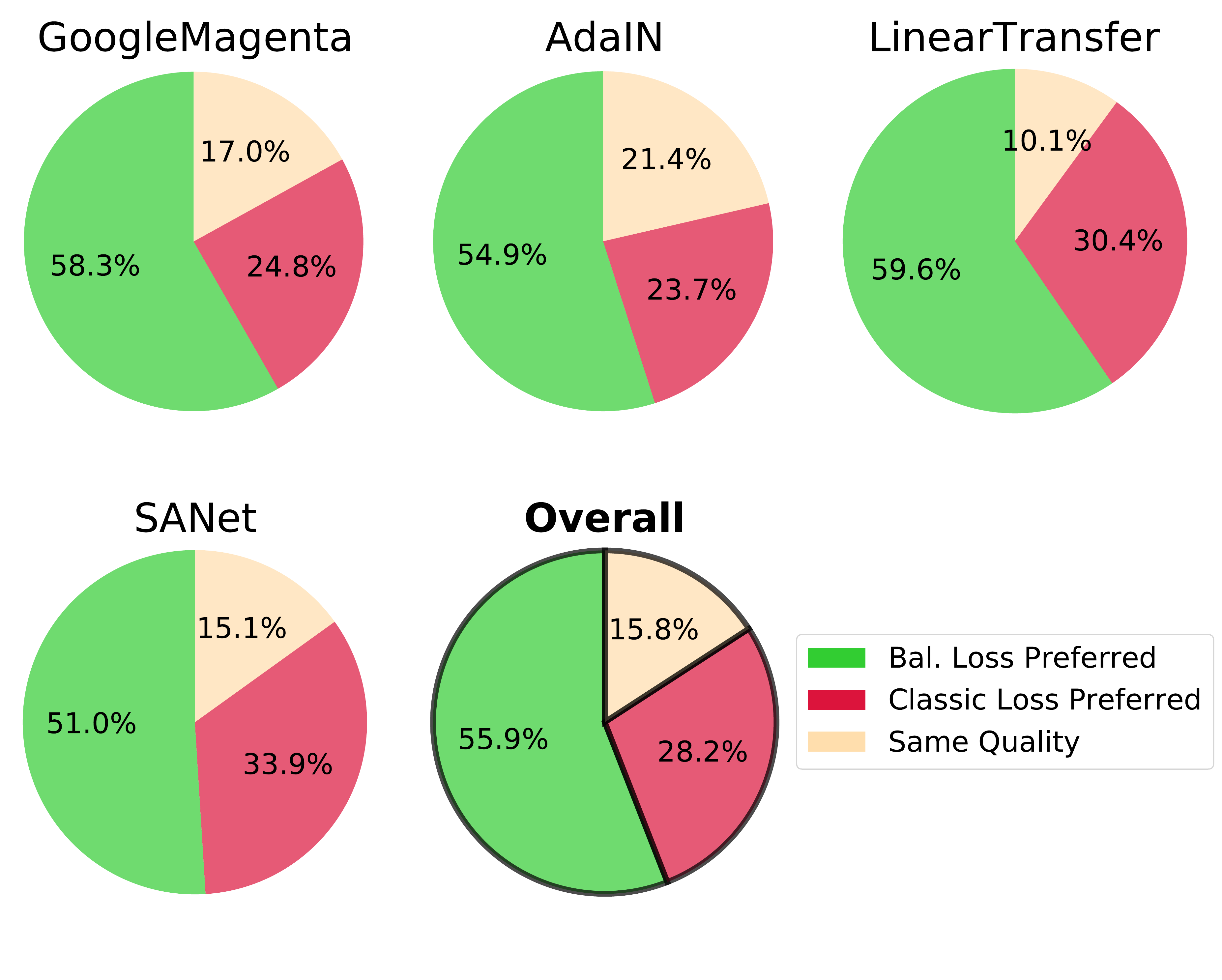}
    \caption{Human evaluation results. On all methods, users prefer images stylized with models trained with our loss.}
    \label{fig.user_study}
    \vspace{-15pt}
\end{figure}

We use the winning solution of the PBN challenge (\hyperlink{https://github.com/inejc/painters}{https://github.com/inejc/painters}) to compute the deception rate. First, we use the model to generate 2,048-dimensional features for all the style images. Next, for each stylized image, we extract its features using the same model and find its nearest style image using $L_2$ distance. For a successful deception, the artist of the nearest neighbor must match that of the style image used for generating the stylized sample.

\Cref{tab.deception_rate} summarizes the results, showing large boosts in deception rate from adopting our new loss. The improvements range from a massive 83\% minimum relative improvement for SANet to 111\% (or 2.1x) for LinearTransfer.

\vspace{5pt}\noindent\textbf{Human Evaluation.}\quad We further quantify improvement in stylization quality due to our new loss through human evaluation. We randomly sample 5,000 content-style pairs with content images from the ImageNet test set and style images each from the test sets of PBN~\cite{pbn} and Describable Textures Dataset (DTD)~\cite{dtd}. For each pair and for each of the four AST methods (\Cref{tab:ast_methods}), we generate two stylized images: one using the model trained with the classic loss and another with that trained using our new loss. Random subsets of images are then voted on by annotators, who are requested to select one of the two stylized images for each pair based on their preference of stylization quality. In total, 2,200 annotations were collected, and show that 28.2\% votes prefer stylized images from models trained with the classic loss while a much larger 55.9\% prefer those trained with our new balanced loss, with 15.8\% finding them of similar quality. \Cref{fig.user_study} provides the breakdown for each model, showing that models trained with our loss generate results that are more preferred by human perception.


\section{Conclusion}
\label{sec.conclusion}
In this work, we systematically studied the discrepancy between the classic AST style loss and human perception of stylization quality. We identified the root cause of the issue as the style-agnostic aggregation of sample-wise losses during training and derived theoretical bounds of the style loss to design a new style-balanced loss with style-aware normalization. We showed that unlike the classic loss, our new loss is positively correlated with human perception. Finally, experimental results show up to $111\%$ and $98\%$ relative improvements in Deception Rate and human preference, respectively. Future work can adopt our new loss in related problems, \eg, video~\cite{LinearTransfer,chen2017coherent,huang2017real,ruder2018artistic,ruder2016artistic,chen2018stereoscopic,gao2020fast,chen2020optical} and photo~\cite{LinearTransfer,yim2020filter,luan2017deep,penhouet2019automated} stylization, texture synthesis~\cite{efros1999texture,efros2001image,kwatra2003graphcut,wei2000fast}, \etc. Future work can also derive tighter bounds for the style loss to improve style-aware normalization.

\clearpage

{\small
\bibliographystyle{ieee_fullname}
\bibliography{egbib}

\begin{thebibliography}{10}\itemsep=-1pt

\bibitem{chen2017coherent}
Dongdong Chen, Jing Liao, Lu Yuan, Nenghai Yu, and Gang Hua.
\newblock Coherent online video style transfer.
\newblock In {\em IEEE International Conference on Computer Vision}, pages
  1105--1114, 2017.

\bibitem{chen2017stylebank}
Dongdong Chen, Lu Yuan, Jing Liao, Nenghai Yu, and Gang Hua.
\newblock Stylebank: An explicit representation for neural image style
  transfer.
\newblock In {\em IEEE conference on Computer Vision and Pattern Recognition},
  pages 1897--1906, 2017.

\bibitem{chen2018stereoscopic}
Dongdong Chen, Lu Yuan, Jing Liao, Nenghai Yu, and Gang Hua.
\newblock Stereoscopic neural style transfer.
\newblock In {\em IEEE Conference on Computer Vision and Pattern Recognition},
  pages 6654--6663, 2018.

\bibitem{chen2016fast}
Tian~Qi Chen and Mark Schmidt.
\newblock Fast patch-based style transfer of arbitrary style.
\newblock In {\em NIPS Workshop on Constructive Machine Learning}, 2016.

\bibitem{chen2020optical}
Xinghao Chen, Yiman Zhang, Yunhe Wang, Han Shu, Chunjing Xu, and Chang Xu.
\newblock Optical flow distillation: Towards efficient and stable video style
  transfer.
\newblock In {\em European Conference on Computer Vision}, pages 614--630.
  Springer, 2020.

\bibitem{chen2018gradnorm}
Zhao Chen, Vijay Badrinarayanan, Chen-Yu Lee, and Andrew Rabinovich.
\newblock Gradnorm: Gradient normalization for adaptive loss balancing in deep
  multitask networks.
\newblock In {\em International Conference on Machine Learning}, pages
  794--803. PMLR, 2018.

\bibitem{dtd}
M. Cimpoi, S. Maji, I. Kokkinos, S. Mohamed, , and A. Vedaldi.
\newblock Describing textures in the wild.
\newblock In {\em IEEE conference on Computer Vision and Pattern Recognition},
  2014.

\bibitem{deng2009imagenet}
Jia Deng, Wei Dong, Richard Socher, Li-Jia Li, Kai Li, and Li Fei-Fei.
\newblock Imagenet: A large-scale hierarchical image database.
\newblock In {\em IEEE conference on Computer Vision and Pattern Recognition},
  pages 248--255. Ieee, 2009.

\bibitem{dumoulin2016learned}
Vincent Dumoulin, Jonathon Shlens, and Manjunath Kudlur.
\newblock A learned representation for artistic style.
\newblock {\em arXiv preprint arXiv:1610.07629}, 2016.

\bibitem{efros2001image}
Alexei~A Efros and William~T Freeman.
\newblock Image quilting for texture synthesis and transfer.
\newblock In {\em Conference on Computer Graphics and Interactive Techniques},
  pages 341--346, 2001.

\bibitem{efros1999texture}
Alexei~A Efros and Thomas~K Leung.
\newblock Texture synthesis by non-parametric sampling.
\newblock In {\em IEEE International Conference on Computer Vision}, volume~2,
  pages 1033--1038. IEEE, 1999.

\bibitem{gao2020fast}
Wei Gao, Yijun Li, Yihang Yin, and Ming-Hsuan Yang.
\newblock Fast video multi-style transfer.
\newblock In {\em IEEE Winter Conference on Applications of Computer Vision},
  pages 3222--3230, 2020.

\bibitem{gatys2016image}
Leon~A Gatys, Alexander~S Ecker, and Matthias Bethge.
\newblock Image style transfer using convolutional neural networks.
\newblock In {\em IEEE conference on Computer Vision and Pattern Recognition},
  pages 2414--2423, 2016.

\bibitem{megenta2}
Golnaz Ghiasi, Honglak Lee, Manjunath Kudlur, Vincent Dumoulin, and Jonathon
  Shlens.
\newblock Exploring the structure of a real-time, arbitrary neural artistic
  stylization network.
\newblock In {\em British Machine Vision Conference}, 2017.

\bibitem{gu2018arbitrary}
Shuyang Gu, Congliang Chen, Jing Liao, and Lu Yuan.
\newblock Arbitrary style transfer with deep feature reshuffle.
\newblock In {\em IEEE Conference on Computer Vision and Pattern Recognition},
  pages 8222--8231, 2018.

\bibitem{guo2018dynamic}
Michelle Guo, Albert Haque, De-An Huang, Serena Yeung, and Li Fei-Fei.
\newblock Dynamic task prioritization for multitask learning.
\newblock In {\em European Conference on Computer Vision}, pages 270--287,
  2018.

\bibitem{hu2020aesthetic}
Zhiyuan Hu, Jia Jia, Bei Liu, Yaohua Bu, and Jianlong Fu.
\newblock Aesthetic-aware image style transfer.
\newblock In {\em ACM International Conference on Multimedia}, pages
  3320--3329, 2020.

\bibitem{huang2017real}
Haozhi Huang, Hao Wang, Wenhan Luo, Lin Ma, Wenhao Jiang, Xiaolong Zhu, Zhifeng
  Li, and Wei Liu.
\newblock Real-time neural style transfer for videos.
\newblock In {\em IEEE Conference on Computer Vision and Pattern Recognition},
  pages 783--791, 2017.

\bibitem{AdaIn}
Xun Huang and Serge Belongie.
\newblock Arbitrary style transfer in real-time with adaptive instance
  normalization.
\newblock In {\em IEEE International Conference on Computer Vision}, pages
  1501--1510, 2017.

\bibitem{jing2020dynamic}
Yongcheng Jing, Xiao Liu, Yukang Ding, Xinchao Wang, Errui Ding, Mingli Song,
  and Shilei Wen.
\newblock Dynamic instance normalization for arbitrary style transfer.
\newblock In {\em AAAI Conference on Artificial Intelligence}, volume~34, pages
  4369--4376, 2020.

\bibitem{johnson2016perceptual}
Justin Johnson, Alexandre Alahi, and Li Fei-Fei.
\newblock Perceptual losses for real-time style transfer and super-resolution.
\newblock In {\em European Conference on Computer Vision}, pages 694--711.
  Springer, 2016.

\bibitem{kendall2018multi}
Alex Kendall, Yarin Gal, and Roberto Cipolla.
\newblock Multi-task learning using uncertainty to weigh losses for scene
  geometry and semantics.
\newblock In {\em IEEE conference on Computer Vision and Pattern Recognition},
  pages 7482--7491, 2018.

\bibitem{kolkin2019style}
Nicholas Kolkin, Jason Salavon, and Gregory Shakhnarovich.
\newblock Style transfer by relaxed optimal transport and self-similarity.
\newblock In {\em IEEE Conference on Computer Vision and Pattern Recognition},
  pages 10051--10060, 2019.

\bibitem{kotovenko2019content2}
Dmytro Kotovenko, Artsiom Sanakoyeu, Sabine Lang, and Bjorn Ommer.
\newblock Content and style disentanglement for artistic style transfer.
\newblock In {\em IEEE International Conference on Computer Vision}, pages
  4422--4431, 2019.

\bibitem{kotovenko2019content}
Dmytro Kotovenko, Artsiom Sanakoyeu, Pingchuan Ma, Sabine Lang, and Bjorn
  Ommer.
\newblock A content transformation block for image style transfer.
\newblock In {\em IEEE Conference on Computer Vision and Pattern Recognition},
  pages 10032--10041, 2019.

\bibitem{kwatra2003graphcut}
Vivek Kwatra, Arno Sch{\"o}dl, Irfan Essa, Greg Turk, and Aaron Bobick.
\newblock Graphcut textures: image and video synthesis using graph cuts.
\newblock {\em ACM Transactions on Graphics}, 22(3):277--286, 2003.

\bibitem{li2016precomputed}
Chuan Li and Michael Wand.
\newblock Precomputed real-time texture synthesis with markovian generative
  adversarial networks.
\newblock In {\em European Conference on Computer Vision}, pages 702--716.
  Springer, 2016.

\bibitem{li2017laplacian}
Shaohua Li, Xinxing Xu, Liqiang Nie, and Tat-Seng Chua.
\newblock Laplacian-steered neural style transfer.
\newblock In {\em ACM International Conference on Multimedia}, pages
  1716--1724, 2017.

\bibitem{LinearTransfer}
Xueting Li, Sifei Liu, Jan Kautz, and Ming-Hsuan Yang.
\newblock Learning linear transformations for fast image and video style
  transfer.
\newblock In {\em IEEE conference on Computer Vision and Pattern Recognition},
  2019.

\bibitem{li2017diversified}
Yijun Li, Chen Fang, Jimei Yang, Zhaowen Wang, Xin Lu, and Ming-Hsuan Yang.
\newblock Diversified texture synthesis with feed-forward networks.
\newblock In {\em IEEE Conference on Computer Vision and Pattern Recognition},
  pages 3920--3928, 2017.

\bibitem{WCT}
Yijun Li, Chen Fang, Jimei Yang, Zhaowen Wang, Xin Lu, and Ming-Hsuan Yang.
\newblock Universal style transfer via feature transforms.
\newblock In {\em Advances in Neural Information Processing Systems}, pages
  386--396, 2017.

\bibitem{li2017demystifying}
Yanghao Li, Naiyan Wang, Jiaying Liu, and Xiaodi Hou.
\newblock Demystifying neural style transfer.
\newblock In {\em International Joint Conference on Artificial Intelligence},
  pages 2230--2236, 2017.

\bibitem{mscoco}
Tsung-Yi Lin, Michael Maire, Serge Belongie, James Hays, Pietro Perona, Deva
  Ramanan, Piotr Doll{\'a}r, and C~Lawrence Zitnick.
\newblock Microsoft coco: Common objects in context.
\newblock In {\em European Conference on Computer Vision}, pages 740--755.
  Springer, 2014.

\bibitem{liu2019end}
Shikun Liu, Edward Johns, and Andrew~J Davison.
\newblock End-to-end multi-task learning with attention.
\newblock In {\em IEEE Conference on Computer Vision and Pattern Recognition},
  pages 1871--1880, 2019.

\bibitem{luan2017deep}
Fujun Luan, Sylvain Paris, Eli Shechtman, and Kavita Bala.
\newblock Deep photo style transfer.
\newblock In {\em IEEE Conference on Computer Vision and Pattern Recognition},
  pages 4990--4998, 2017.

\bibitem{pbn}
K Nichol.
\newblock {\em Painter by numbers}, 2016.

\bibitem{SANet}
Dae~Young Park and Kwang~Hee Lee.
\newblock Arbitrary style transfer with style-attentional networks.
\newblock In {\em IEEE Conference on Computer Vision and Pattern Recognition},
  pages 5880--5888, 2019.

\bibitem{penhouet2019automated}
Sebastian Penhou{\"e}t and Paul Sanzenbacher.
\newblock Automated deep photo style transfer.
\newblock {\em arXiv preprint arXiv:1901.03915}, 2019.

\bibitem{risser2017stable}
Eric Risser, Pierre Wilmot, and Connelly Barnes.
\newblock Stable and controllable neural texture synthesis and style transfer
  using histogram losses.
\newblock {\em arXiv preprint arXiv:1701.08893}, 2017.

\bibitem{ruder2016artistic}
Manuel Ruder, Alexey Dosovitskiy, and Thomas Brox.
\newblock Artistic style transfer for videos.
\newblock In {\em German Conference on Pattern Recognition}, pages 26--36.
  Springer, 2016.

\bibitem{ruder2018artistic}
Manuel Ruder, Alexey Dosovitskiy, and Thomas Brox.
\newblock Artistic style transfer for videos and spherical images.
\newblock {\em International Journal of Computer Vision}, 126(11):1199--1219,
  2018.

\bibitem{sanakoyeu2018style}
Artsiom Sanakoyeu, Dmytro Kotovenko, Sabine Lang, and Bjorn Ommer.
\newblock A style-aware content loss for real-time hd style transfer.
\newblock In {\em European Conference on Computer Vision}, pages 698--714,
  2018.

\bibitem{shen2018neural}
Falong Shen, Shuicheng Yan, and Gang Zeng.
\newblock Neural style transfer via meta networks.
\newblock In {\em IEEE Conference on Computer Vision and Pattern Recognition},
  pages 8061--8069, 2018.

\bibitem{sheng2018avatar}
Lu Sheng, Ziyi Lin, Jing Shao, and Xiaogang Wang.
\newblock Avatar-net: Multi-scale zero-shot style transfer by feature
  decoration.
\newblock In {\em IEEE Conference on Computer Vision and Pattern Recognition},
  pages 8242--8250, 2018.

\bibitem{vgg}
Karen Simonyan and Andrew Zisserman.
\newblock Very deep convolutional networks for large-scale image recognition.
\newblock {\em arXiv preprint arXiv:1409.1556}, 2014.

\bibitem{svoboda2020two}
Jan Svoboda, Asha Anoosheh, Christian Osendorfer, and Jonathan Masci.
\newblock Two-stage peer-regularized feature recombination for arbitrary image
  style transfer.
\newblock In {\em IEEE/CVF Conference on Computer Vision and Pattern
  Recognition}, pages 13816--13825, 2020.

\bibitem{ulyanov2016texture}
Dmitry Ulyanov, Vadim Lebedev, Andrea Vedaldi, and Victor~S Lempitsky.
\newblock Texture networks: Feed-forward synthesis of textures and stylized
  images.
\newblock In {\em International Conference on Machine Learning}, volume~1,
  page~4, 2016.

\bibitem{ulyanov2017improved}
Dmitry Ulyanov, Andrea Vedaldi, and Victor Lempitsky.
\newblock Improved texture networks: Maximizing quality and diversity in
  feed-forward stylization and texture synthesis.
\newblock In {\em IEEE Conference on Computer Vision and Pattern Recognition},
  pages 6924--6932, 2017.

\bibitem{wang2020collaborative}
Huan Wang, Yijun Li, Yuehai Wang, Haoji Hu, and Ming-Hsuan Yang.
\newblock Collaborative distillation for ultra-resolution universal style
  transfer.
\newblock In {\em IEEE/CVF Conference on Computer Vision and Pattern
  Recognition}, pages 1860--1869, 2020.

\bibitem{wang2020diversified}
Zhizhong Wang, Lei Zhao, Haibo Chen, Lihong Qiu, Qihang Mo, Sihuan Lin, Wei
  Xing, and Dongming Lu.
\newblock Diversified arbitrary style transfer via deep feature perturbation.
\newblock In {\em IEEE/CVF Conference on Computer Vision and Pattern
  Recognition}, pages 7789--7798, 2020.

\bibitem{wei2000fast}
Li-Yi Wei and Marc Levoy.
\newblock Fast texture synthesis using tree-structured vector quantization.
\newblock In {\em Conference on Computer Graphics and Interactive Techniques},
  pages 479--488, 2000.

\bibitem{yao2019attention}
Yuan Yao, Jianqiang Ren, Xuansong Xie, Weidong Liu, Yong-Jin Liu, and Jun Wang.
\newblock Attention-aware multi-stroke style transfer.
\newblock In {\em IEEE Conference on Computer Vision and Pattern Recognition},
  pages 1467--1475, 2019.

\bibitem{yim2020filter}
Jonghwa Yim, Jisung Yoo, Won-joon Do, Beomsu Kim, and Jihwan Choe.
\newblock Filter style transfer between photos.
\newblock In {\em European Conference on Computer Vision}, pages 103--119.
  Springer, 2020.

\bibitem{yoo2019photorealistic}
Jaejun Yoo, Youngjung Uh, Sanghyuk Chun, Byeongkyu Kang, and Jung-Woo Ha.
\newblock Photorealistic style transfer via wavelet transforms.
\newblock In {\em IEEE International Conference on Computer Vision}, pages
  9036--9045, 2019.

\bibitem{zhang2018multi}
Hang Zhang and Kristin Dana.
\newblock Multi-style generative network for real-time transfer.
\newblock In {\em European Conference on Computer Vision}, pages 0--0, 2018.

\end{thebibliography}
}

\clearpage
\appendix

\twocolumn[{%
\renewcommand\twocolumn[1][]{#1}%
\begin{center}
\textbf{\Large Style-Aware Normalized Loss for Improving Arbitrary Style Transfer}

\vspace{10pt}
\textbf{\large Supplementary Material}
\vspace{10pt}
\end{center}%
}]

\section{Distribution of VGG-19-based Style Losses}

\begin{figure*}[!h]
    \vspace{20pt}
    \centering
    \includegraphics[width=\linewidth]{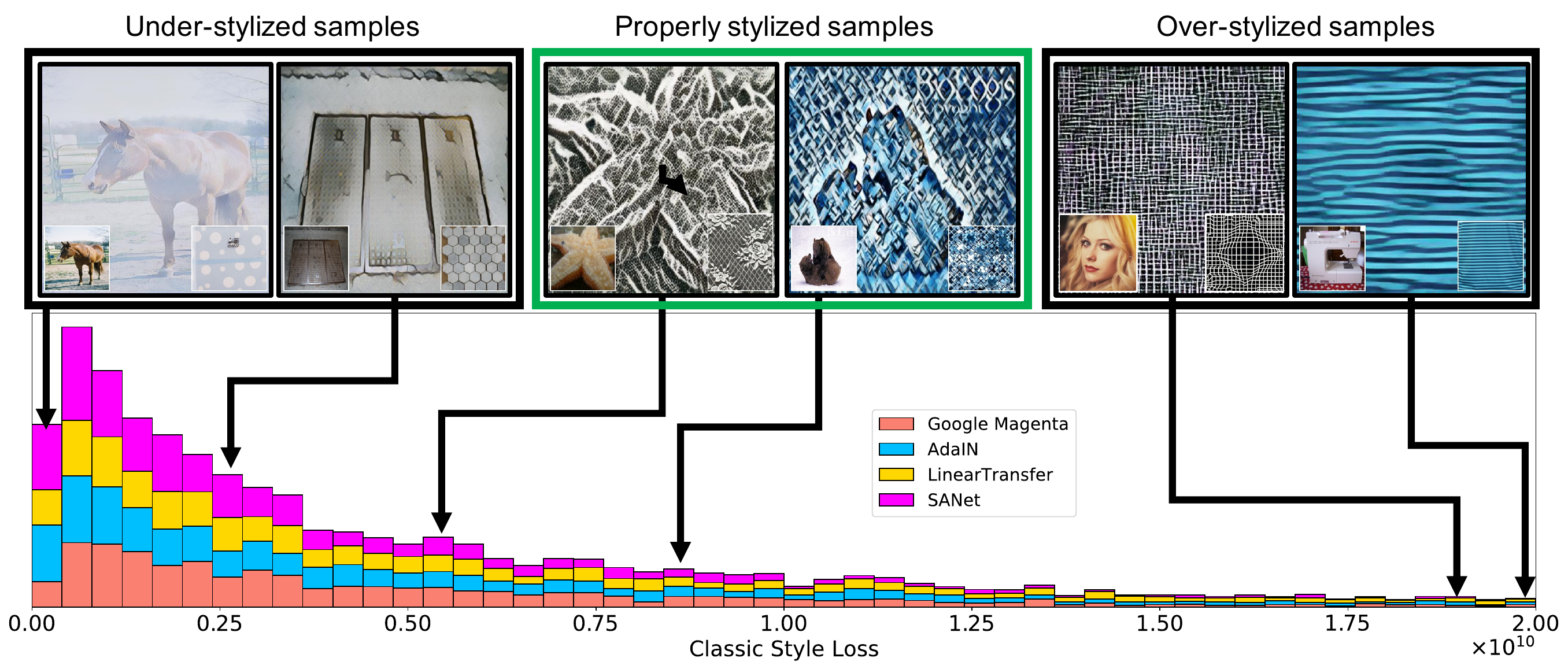}
    \includegraphics[width=\linewidth]{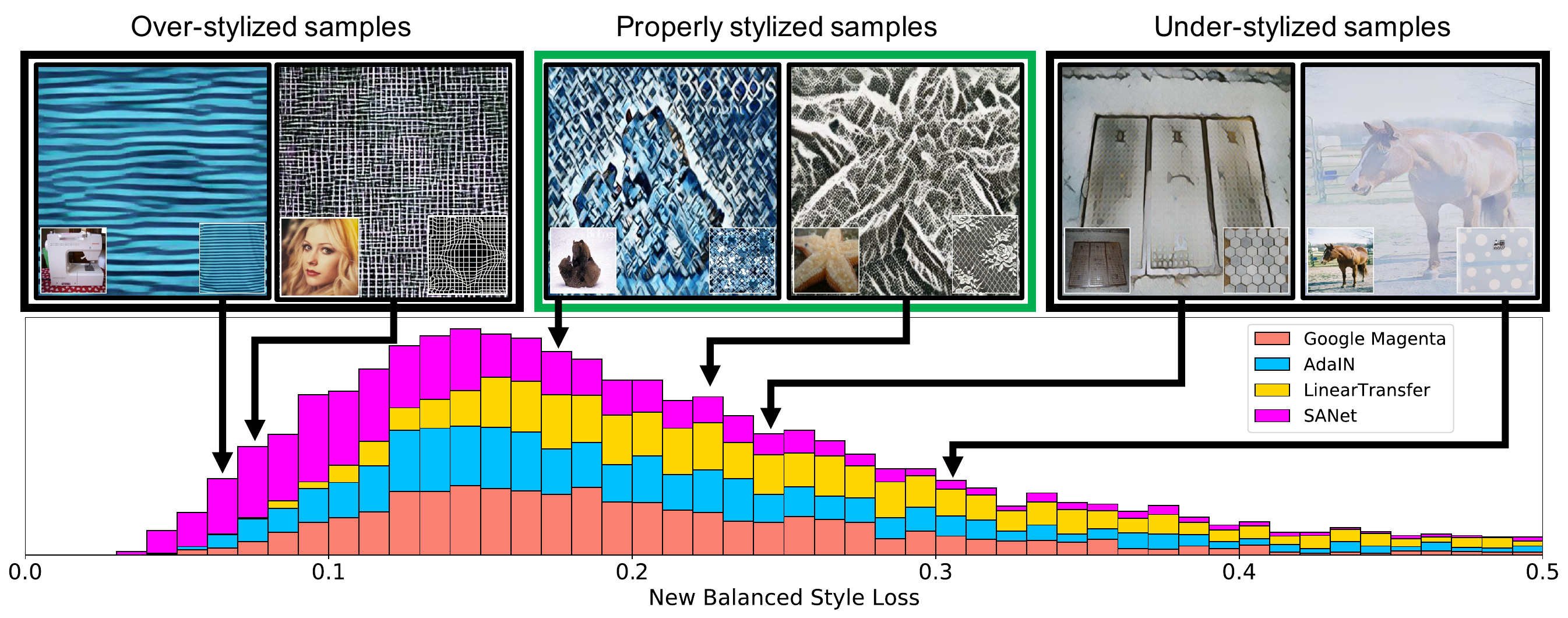}
    \captionof{figure}{Distribution of the classic Gram matrix-based style losses (above) and balanced losses (below) for the AST methods studied in this work. The distribution of VGG-19-based losses does not reflect stylization quality, same as our previous observation with VGG-16.}
    \label{fig.vgg19_loss_hist}
\end{figure*}

We duplicate Study I in Section~\textcolor{blue}{3.2} with VGG-19\cite{vgg} as the loss model. Results are presented in \Cref{fig.vgg19_loss_hist}. As evident, similar conclusion as before can be drawn that the classic style loss does not reflect stylization quality.

\section{Derivations}
In this section, we present derivations of the upper and lower bounds of the classic AST layerwise style loss ( \Cref{eq.upperbound} and \Cref{eq.lowerbound} in the main paper) as discussed in Section~\textcolor{blue}{4.2} and restated below.
\begin{equation}
   \!\! \sup\{\loss_{\textrm{AST}_s}^l(S,P)\} \!\!=\!\! {\|\gram\circ\feat^l(S)\|^2 + \|\gram\circ\feat^l(P)\|^2 \over N^l} \nonumber
\end{equation}
\begin{equation}
   \!\! \inf\{\loss_{\textrm{AST}_s}^l(S,P)\} \!\!=\! \!{(\|\gram\circ\feat^l(S)\| - \|\gram\circ\feat^l(P)\|)^2 \over N^l} \nonumber
\end{equation}
\textbf{Notation.}\quad For brevity, we use $\gram_S$ to represent $\gram\circ\feat^l(S)$ and $\gram_P$ to represent $\gram\circ\feat^l(P)$. $\mathcal{G}_{S,k}$ ($k$=[1, 2, ... $N^l$]) denotes the $k$th element in the Gram matrix $\mathcal{G}_{S}$. We use $\odot$ as the element-wise product between two matrices. $N^l$ is a constant that is equal to the product of spatial dimensions of the feature tensor at layer $l$. 
\begin{proof} The style loss at layer $l$ can be expanded as:
    \begin{align}
        \loss_{\textrm{AST}_s}^l(S,P) &= \frac{\left\| \mathcal{G}_S - \mathcal{G}_P \right \|^2}{N^l} \nonumber \\ 
        &= \frac{1}{N^l} (\left \| \mathcal{G}_S \right \|^2 + \left \| \mathcal{G}_P \right \|^2 - 2 \times (\mathcal{G}_S \odot \mathcal{G}_P)) \nonumber
    \end{align}
    \textbf{Upper bound:} We show that term $\mathcal{G}_S \odot \mathcal{G}_P \geq 0$ by writing it into summation format.
    \begin{equation}
        \mathcal{G}_S \odot \mathcal{G}_P = \sum_{k=1}^{N^l} \mathcal{G}_{S,k} \times \mathcal{G}_{P,k}\geq 0 \nonumber
    \end{equation}
    This is because Gram matrix is computed from features output by a ReLU layer, as a result, all values in the matrix are non-negative. The same conclusion can be drawn for other non-negative activation functions, such as sigmoid and softmax, are used.  Hence, we have
    \begin{equation}
        \loss_{\textrm{AST}_s}^l(S,P) \leq \frac{\left \| \mathcal{G}_S \right \|^2 + \left \| \mathcal{G}_P \right \|^2}{N^l} \nonumber
    \end{equation}
    \textbf{Lower bound:} We show that term $\mathcal{G}_S \odot \mathcal{G}_P \leq \left \| \mathcal{G}_S \right \| \left \| \mathcal{G}_P \right \|$ by using Cauchy-Schwarz inequality.
    \begin{align}
        (\mathcal{G}_S \odot \mathcal{G}_P) &= \sum_{k=1}^{N^l}  \mathcal{G}_{S,k} \times \mathcal{G}_{P,k} \\ \nonumber
        &\stackrel{\textcircled{\raisebox{-0.9pt}{1}}}{=} \mathcal{G}_{S,flatten}^\intercal \cdot \mathcal{G}_{P,flatten} \\ \nonumber
        &\stackrel{\textcircled{\raisebox{-0.9pt}{2}}}{\leq} \left \| \mathcal{G}_S \right \| \left \| \mathcal{G}_P \right \| \\ \nonumber
    \end{align}
    Step $\textcircled{\raisebox{-0.9pt}{1}}$ is by rewriting summation into a dot product. Step $\textcircled{\raisebox{-0.9pt}{2}}$ is because of Cauchy–Schwarz inequality. 
    \begin{align}
        \loss_{\textrm{AST}_s}^l(S,P) &\geq \frac{1}{N^l} (\left \| \mathcal{G}_S \right \|^2 + \left \| \mathcal{G}_P \right \|^2 - 2 \times \left \| \mathcal{G}_S \right \| \left \| \mathcal{G}_P \right \|) \nonumber \\
        &= \frac{1}{N^l}(\left \| \mathcal{G}_S \right \| - \left \| \mathcal{G}_P \right \|)^2 \nonumber
    \end{align}
\end{proof}

\section{Zoomed In Qualitative Results}
\begin{figure*}
    \centering
    
    \begin{tabular}{p{4cm}p{4cm}p{4cm}p{4cm}}
        \centering{Content} & \centering{Style} & \centering{Our Balanced Style Loss} & \centering{Classic Style Loss}
    \end{tabular}
    \rule{\textwidth}{0.2pt}
    Google Magenta
    \includegraphics[width=\linewidth,trim=0 2cm 0 3cm,clip]{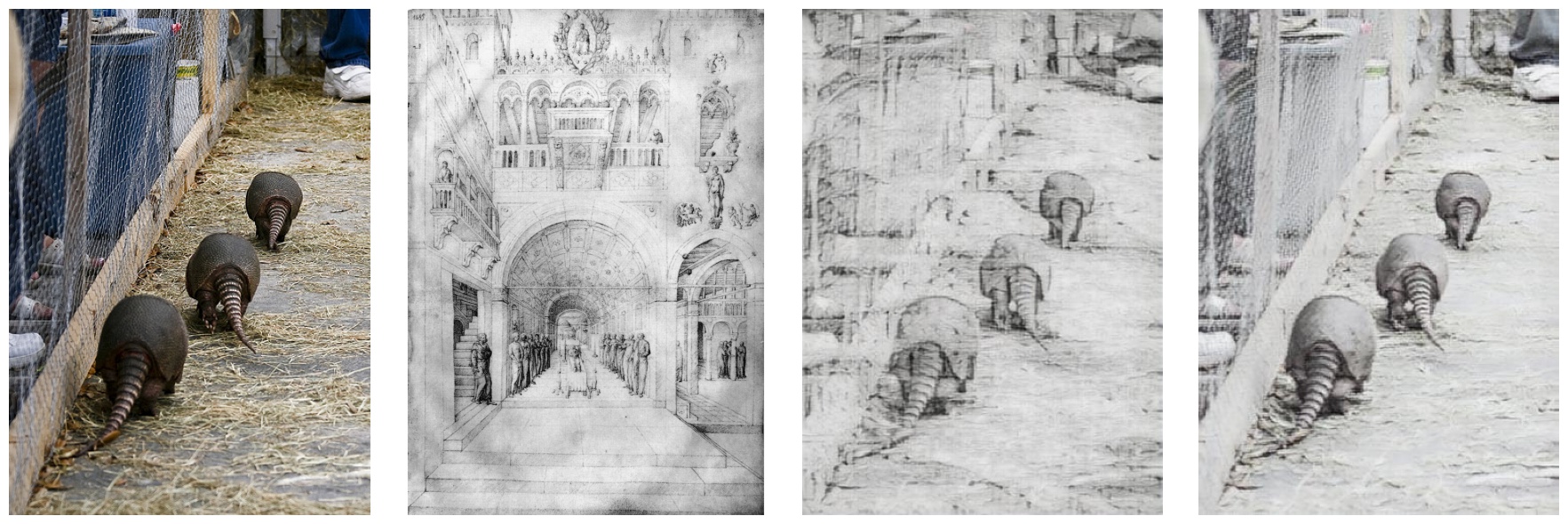}
    \includegraphics[width=\linewidth,trim=0 2cm 0 2cm,clip]{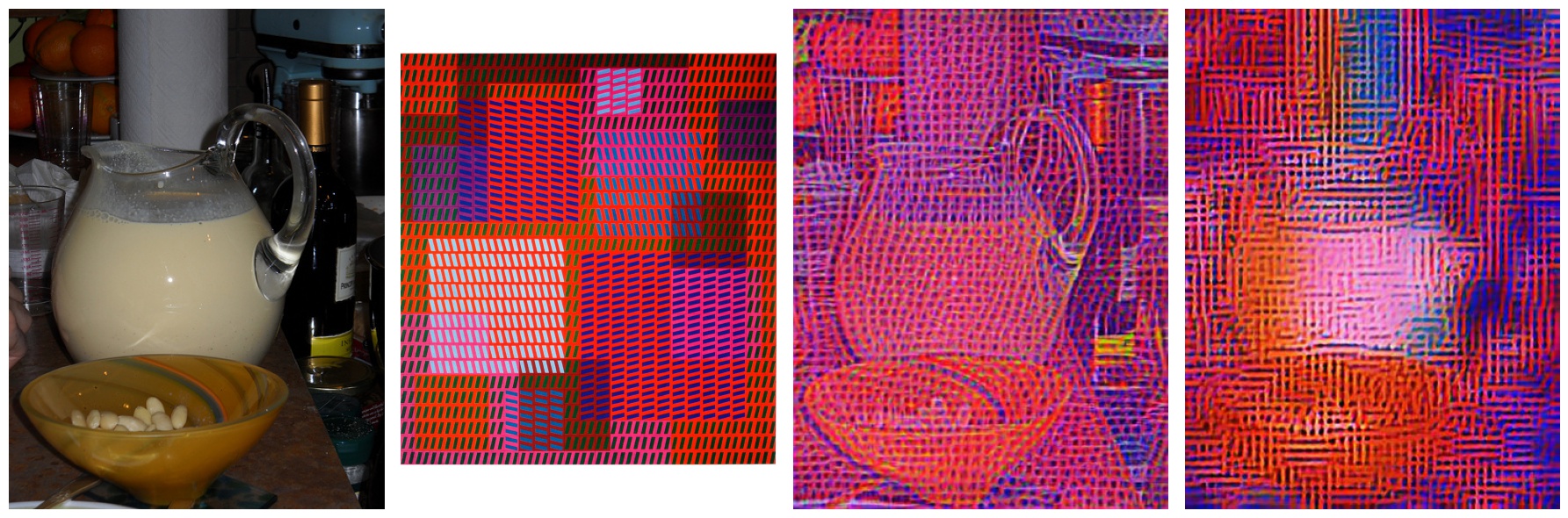}
    \includegraphics[width=\linewidth,trim=0 0.2cm 0 0.2cm,clip]{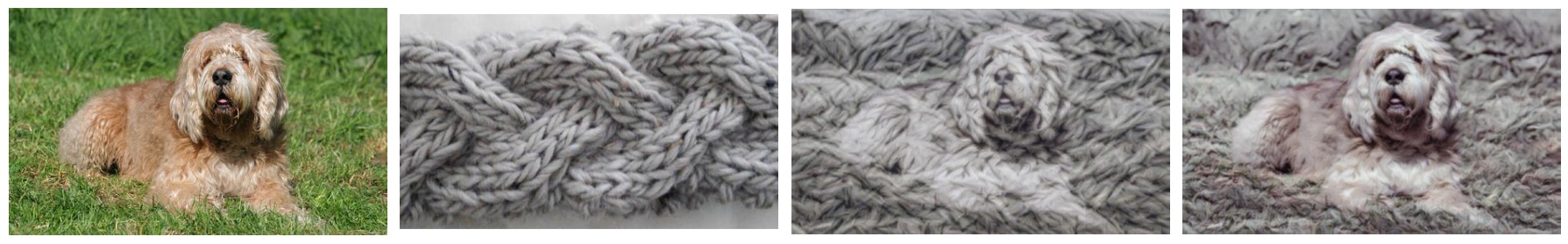}
    \rule{\textwidth}{0.2pt}
    AdaIN
    \includegraphics[width=\linewidth,trim=0 0.8cm 0 0.8cm,clip]{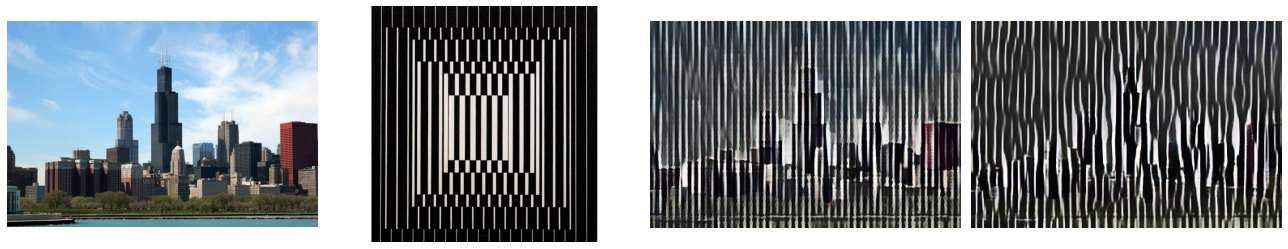}
    \includegraphics[width=\linewidth,trim=0 0cm 0 0cm,clip]{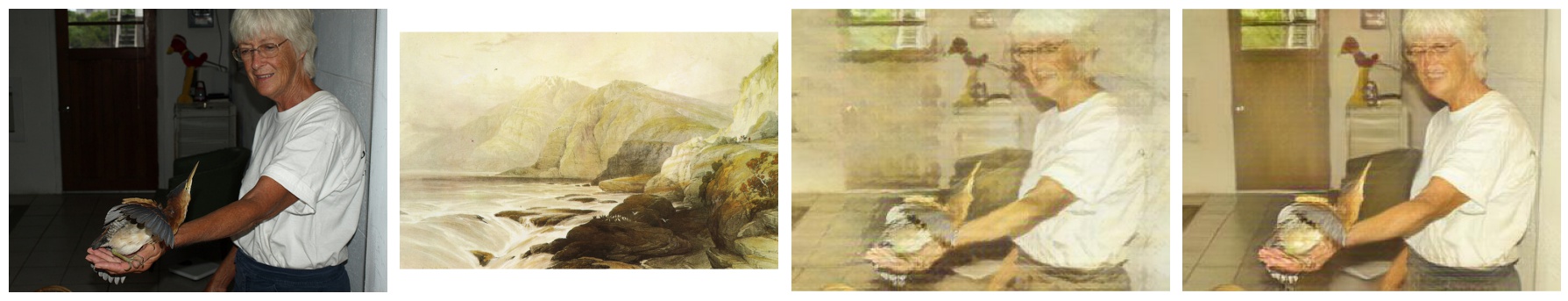}
    \includegraphics[width=\linewidth,trim=0 0cm 0 0cm,clip]{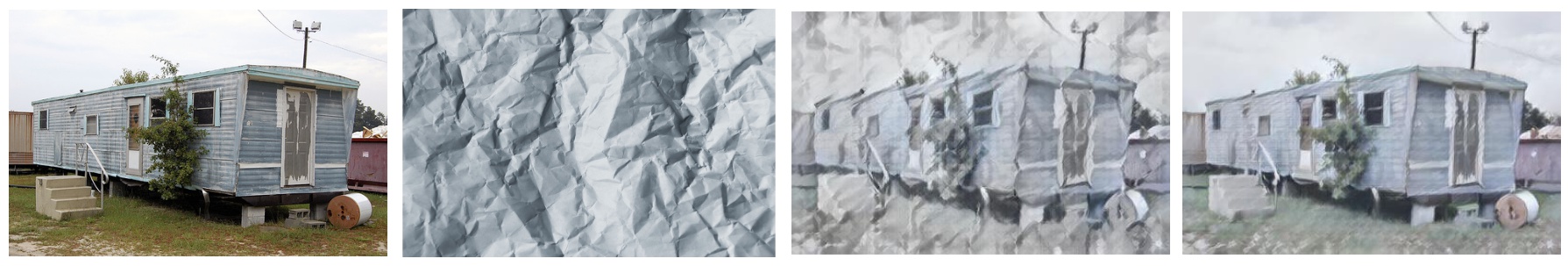}
    \caption{Zoomed in Google Magenta\cite{megenta2} and AdaIN\cite{AdaIn} qualitative samples used in Figure \textcolor{blue}{6}}
    \label{fig.zoom_in_magenta_and_adain}
\end{figure*}

\begin{figure*}
    \centering
    
    \begin{tabular}{p{4cm}p{4cm}p{4cm}p{4cm}}
        \centering{Content} & \centering{Style} & \centering{Our Balanced Style Loss} & \centering{Classic Style Loss}
    \end{tabular}
    \rule{\textwidth}{0.2pt}
    LinearTransfer
    \includegraphics[width=\linewidth,trim=0 0cm 0 4cm,clip]{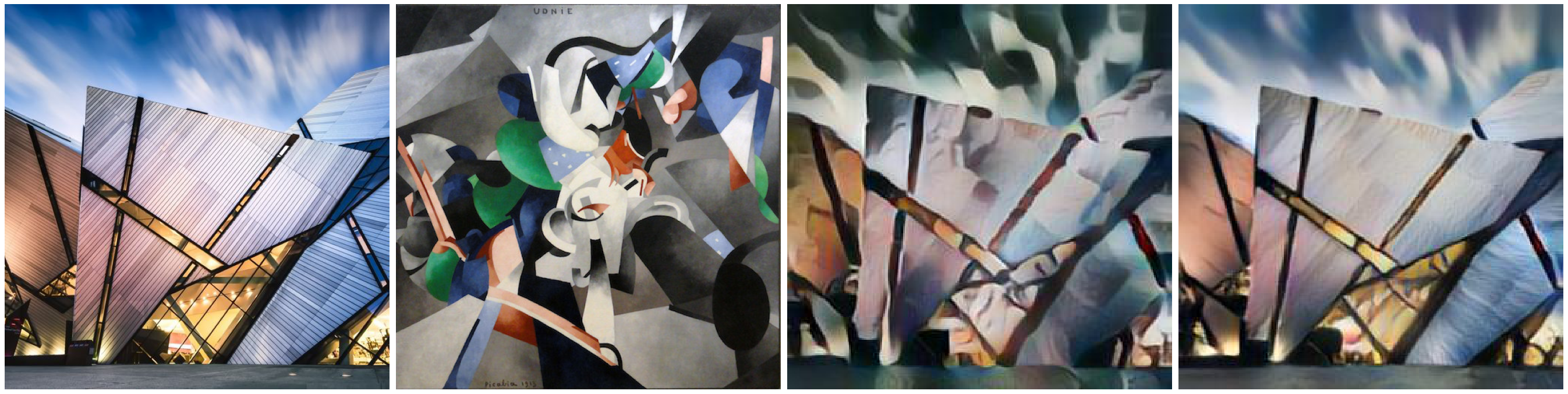}
    \includegraphics[width=\linewidth,trim=0 1cm 0 1cm,clip]{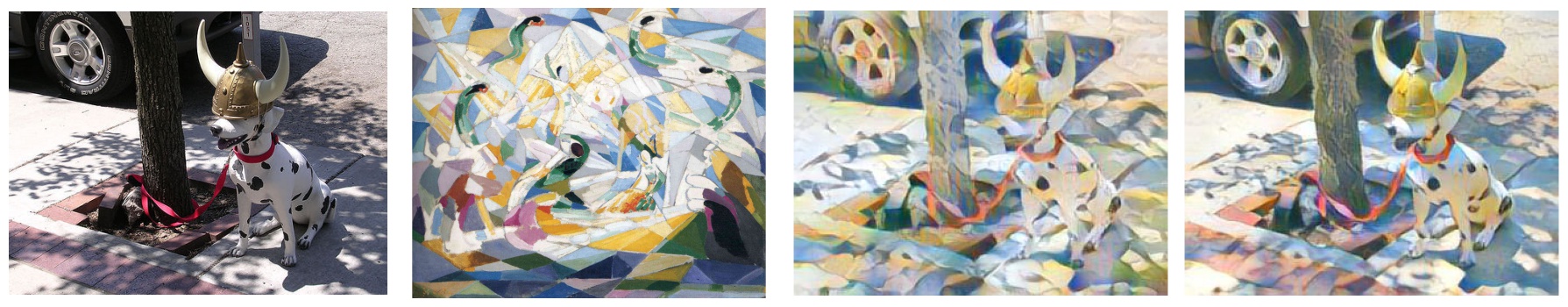}
    \includegraphics[width=\linewidth,trim=0 1cm 0 1cm,clip]{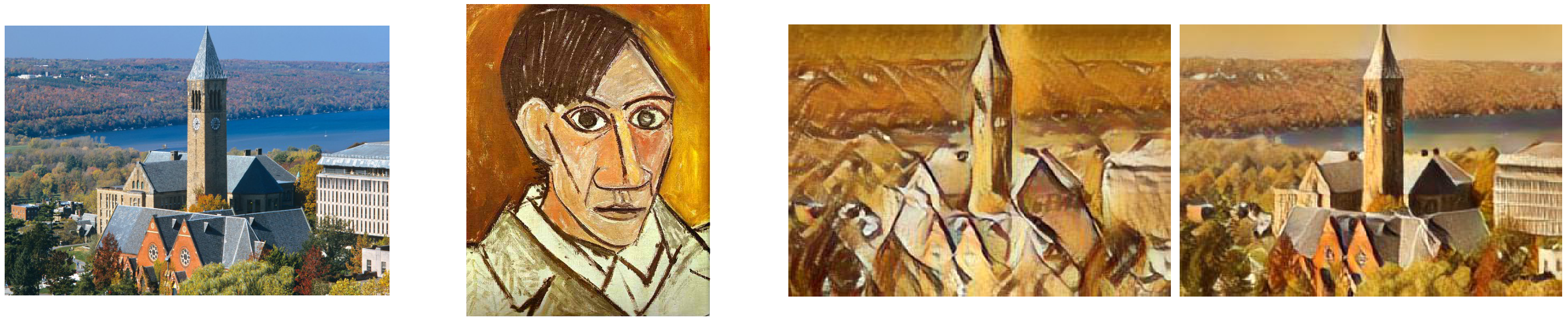}
    \rule{\textwidth}{0.2pt}
    
    SANet
    \includegraphics[width=\linewidth,trim=0 1cm 0 1cm,clip]{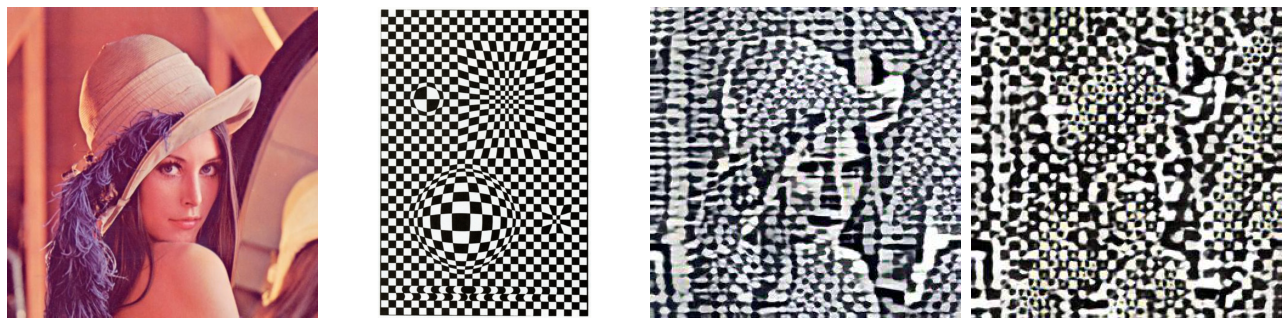}
    \includegraphics[width=\linewidth,trim=0 1cm 0 1cm,clip]{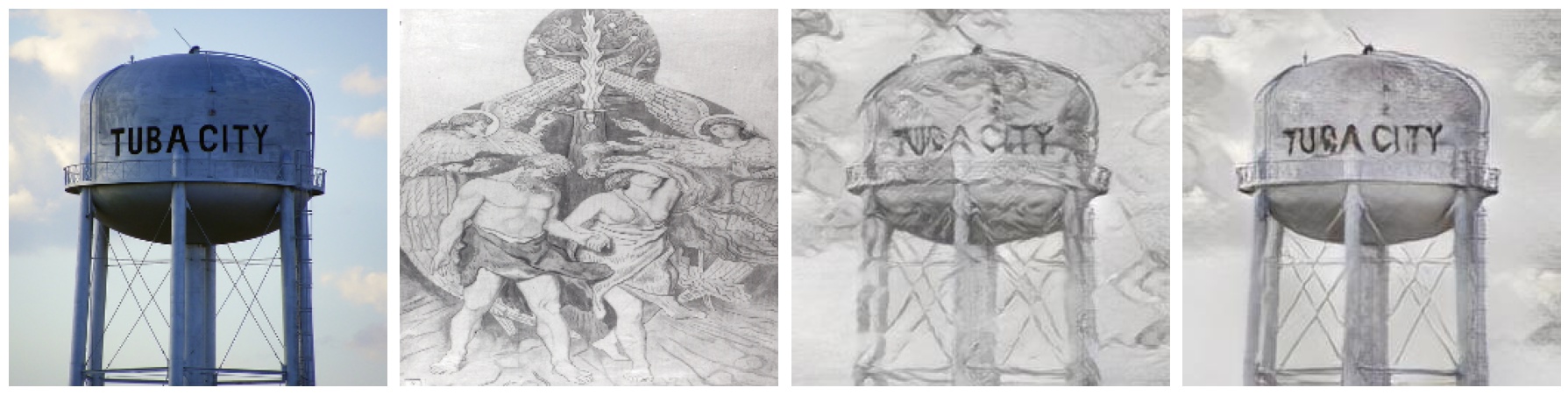}
    \includegraphics[width=\linewidth,trim=0 1cm 0 1cm,clip]{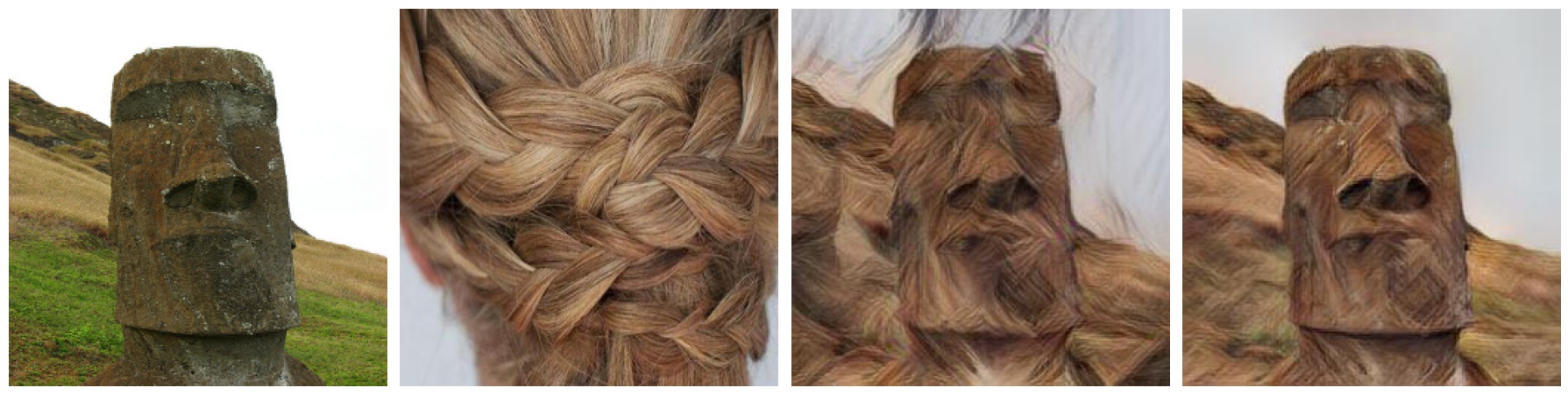}
    \caption{Zoomed in Linear Transfer\cite{LinearTransfer} and SANet\cite{SANet} qualitative samples used in Figure \textcolor{blue}{6}}
    \label{fig.zoom_in_LinearTransfer_and_SANET}
\end{figure*}

We present the zoomed in qualitative results used in Figure \textcolor{blue}{6} in \Cref{fig.zoom_in_magenta_and_adain,fig.zoom_in_LinearTransfer_and_SANET}

\section{Additional Qualitative Results}
\begin{figure*}
    \centering
    \begin{tabular}{p{4cm}p{4cm}p{4cm}p{4cm}}
        \centering{Content} & \centering{Style} & \centering{Our Balanced Style Loss} & \centering{Classic Style Loss}
    \end{tabular}
    
    \includegraphics[width=\linewidth,trim=0 0.8cm 0 0.8cm,clip]{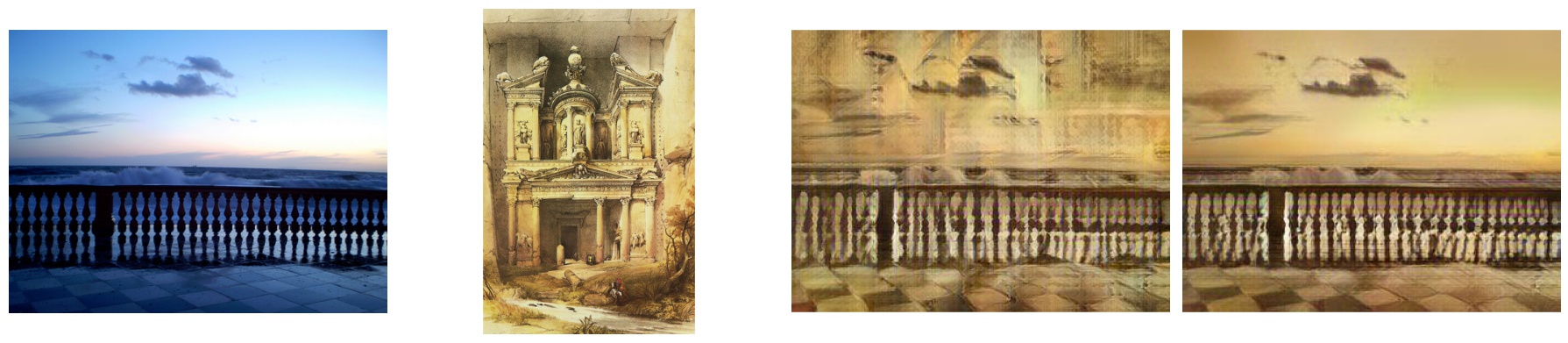}
    \includegraphics[width=\linewidth,trim=0 3cm 0 1cm,clip]{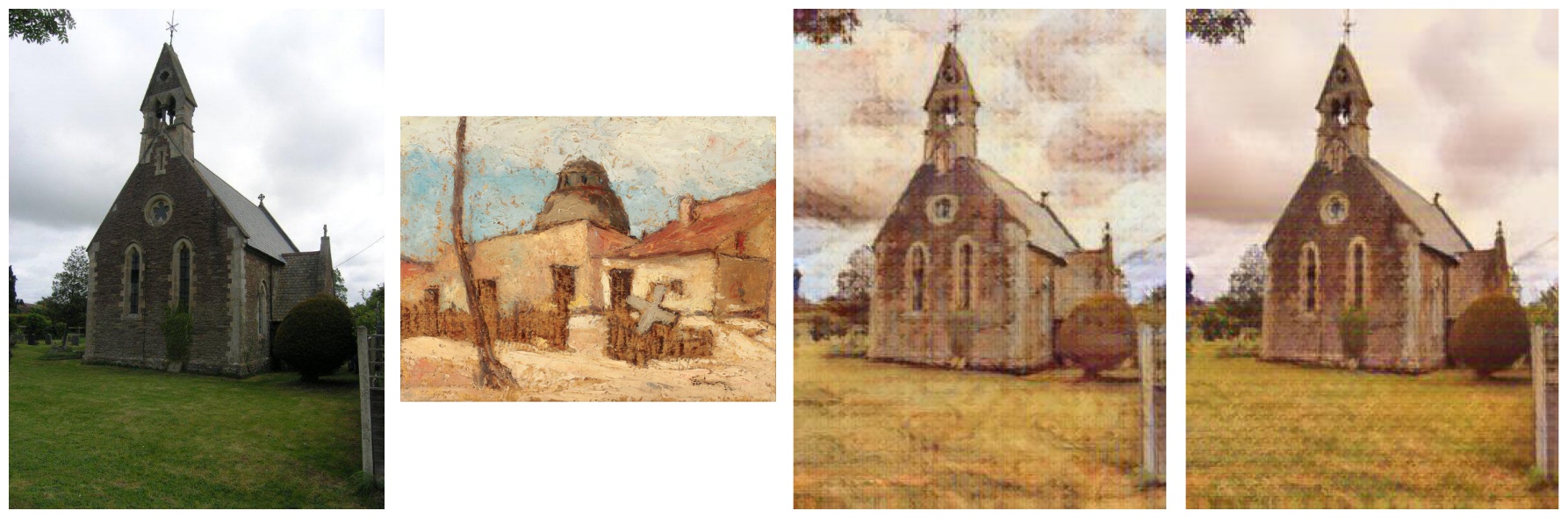}
    \includegraphics[width=\linewidth,trim=0 2cm 0 1.6cm,clip]{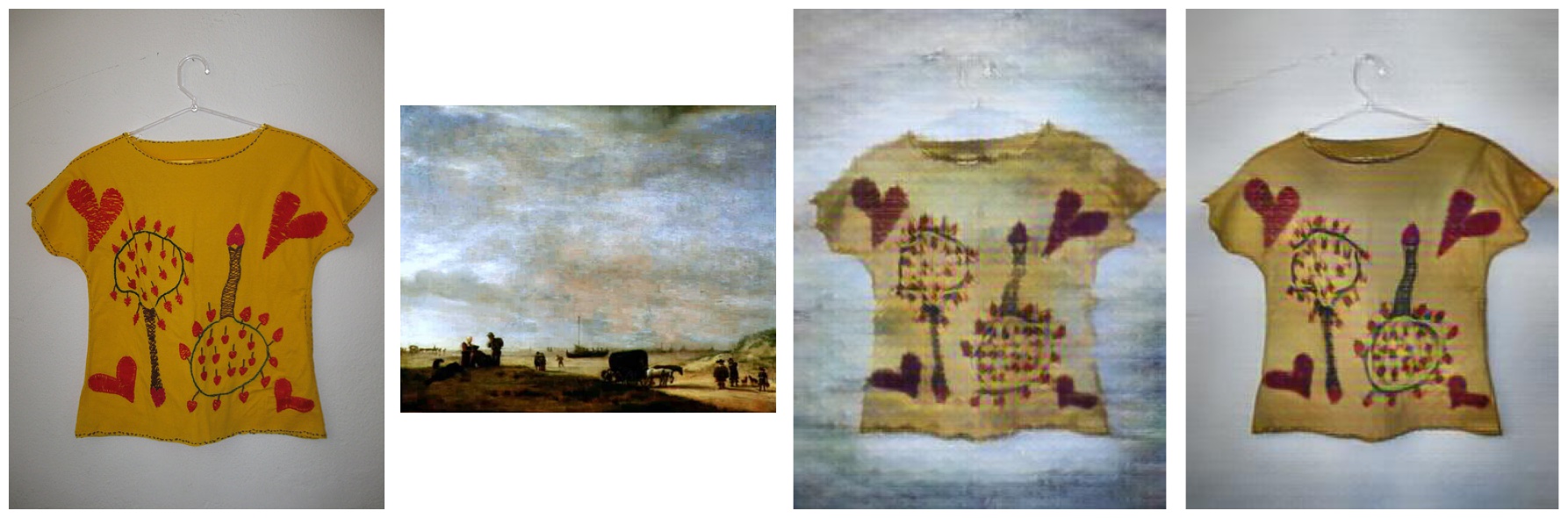}
    \includegraphics[width=\linewidth,trim=0 0.8cm 0 0.8cm,clip]{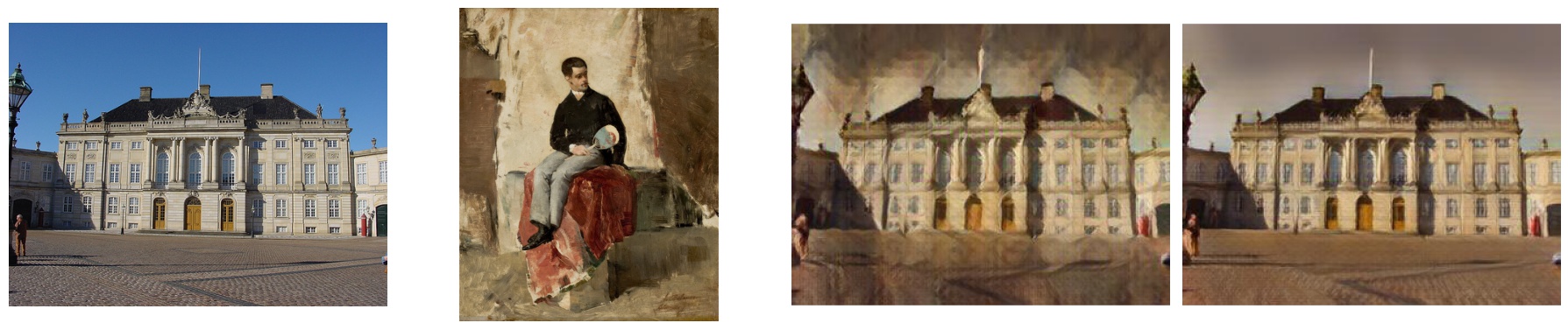}
    \includegraphics[width=\linewidth,trim=0 0.6cm 0 0.6cm,clip]{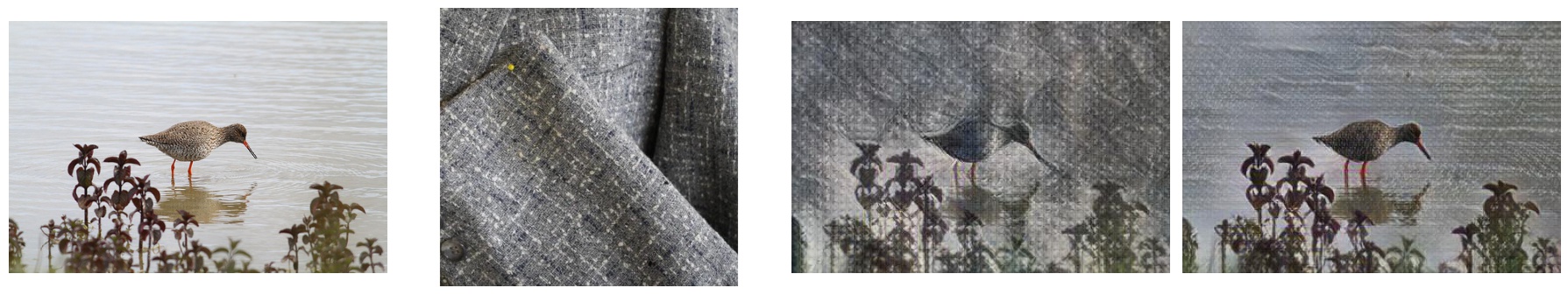}
    \includegraphics[width=\linewidth,trim=0 0.6cm 0 0.6cm,clip]{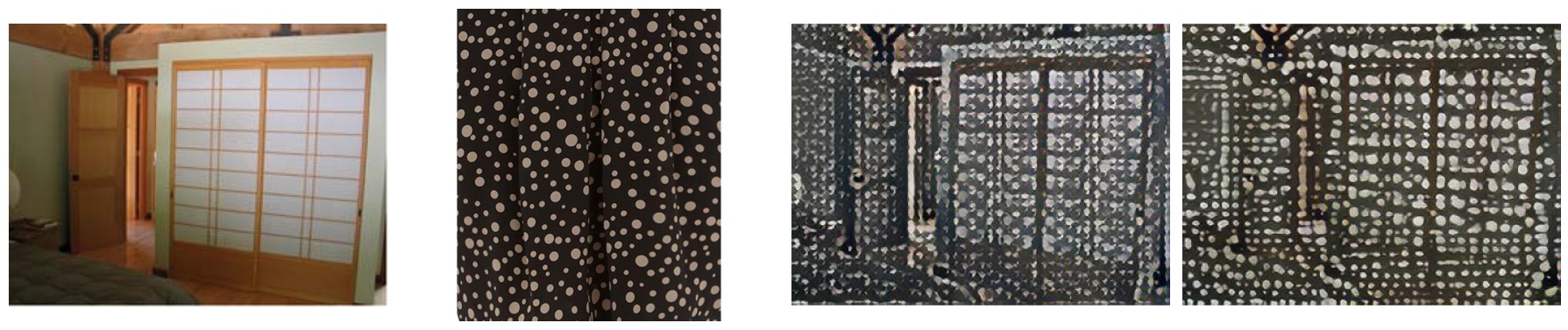}
    \caption{Additional qualitative results for AdaIN\cite{AdaIn}}
    \label{fig.appendix_adain}
\end{figure*}

\begin{figure*}
    \centering
    \begin{tabular}{p{4cm}p{4cm}p{4cm}p{4cm}}
        \centering{Content} & \centering{Style} & \centering{Our Balanced Style Loss} & \centering{Classic Style Loss}
    \end{tabular}
    
    \includegraphics[width=\linewidth,trim=0 0.8cm 0 0.8cm,clip]{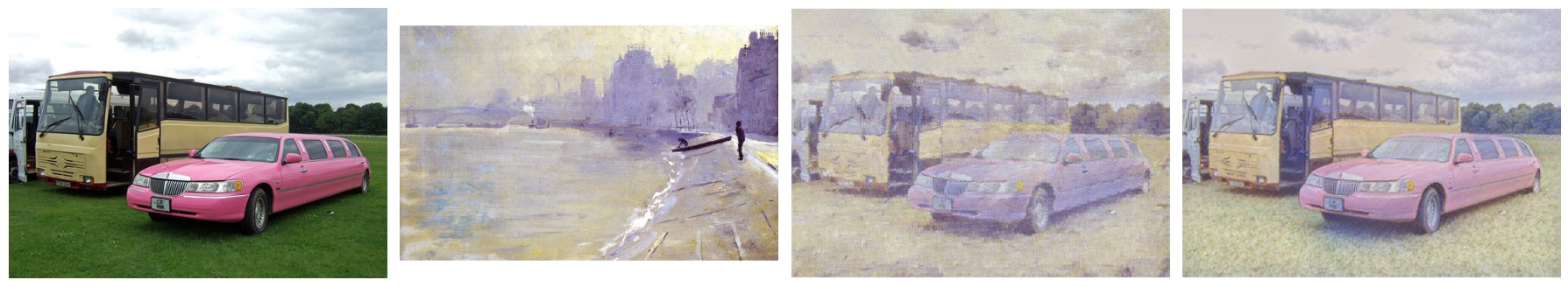}
    \includegraphics[width=\linewidth,trim=0 0cm 0 0cm,clip]{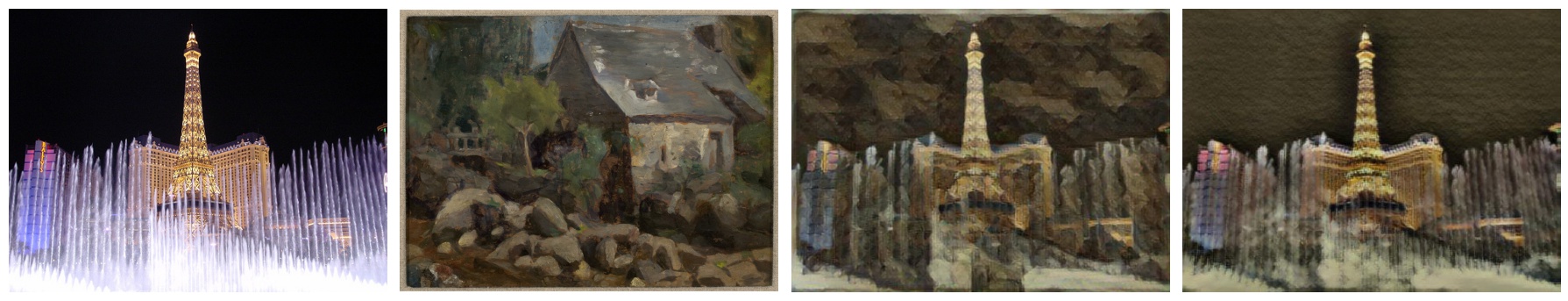}
    \includegraphics[width=\linewidth,trim=0 2cm 0 1cm,clip]{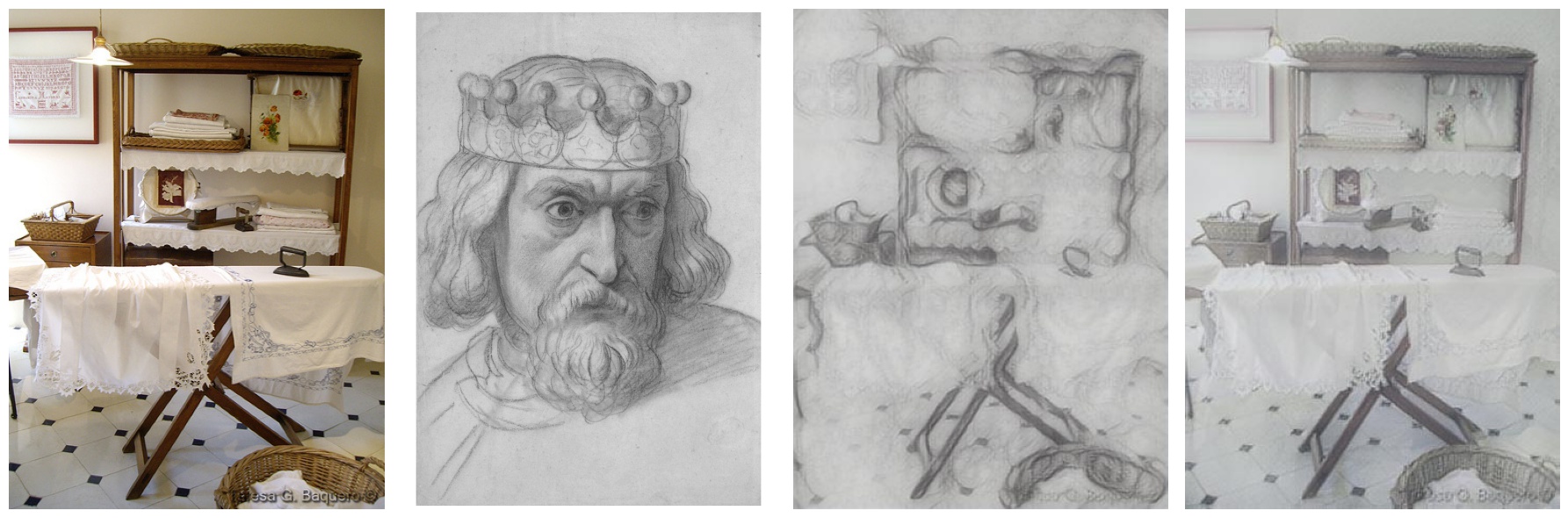}
    \includegraphics[width=\linewidth,trim=0 3cm 0 3cm,clip]{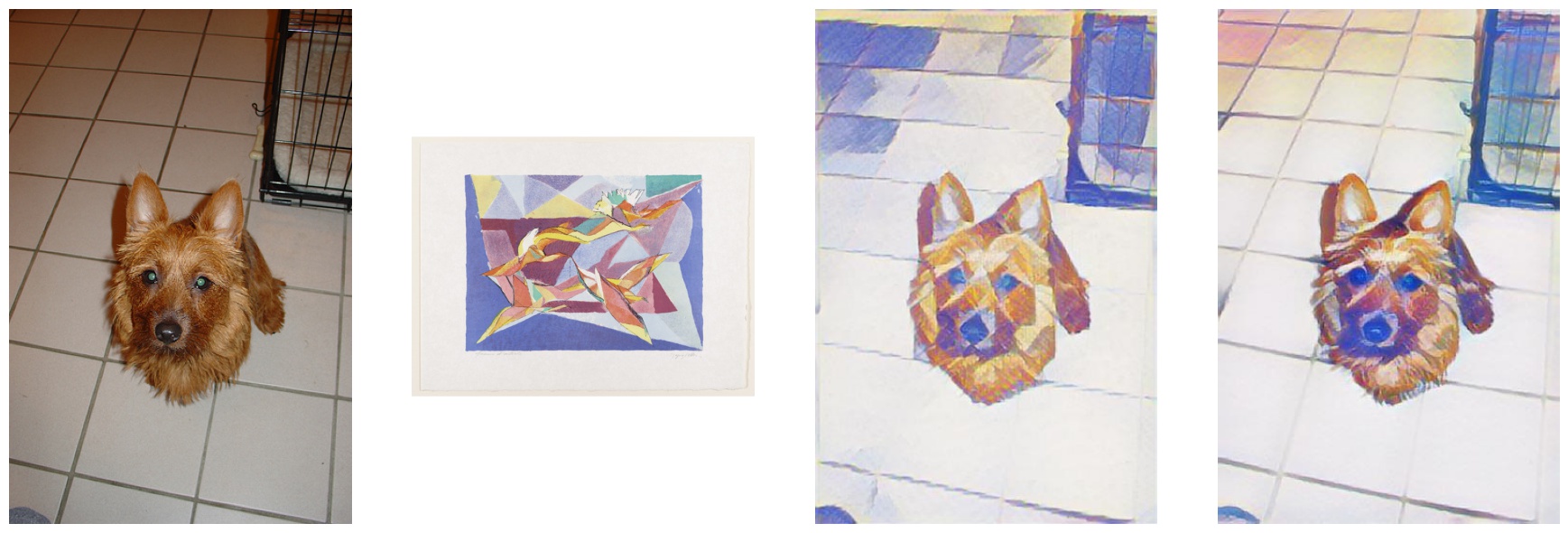}
    \includegraphics[width=\linewidth,trim=0 2cm 0 2cm,clip]{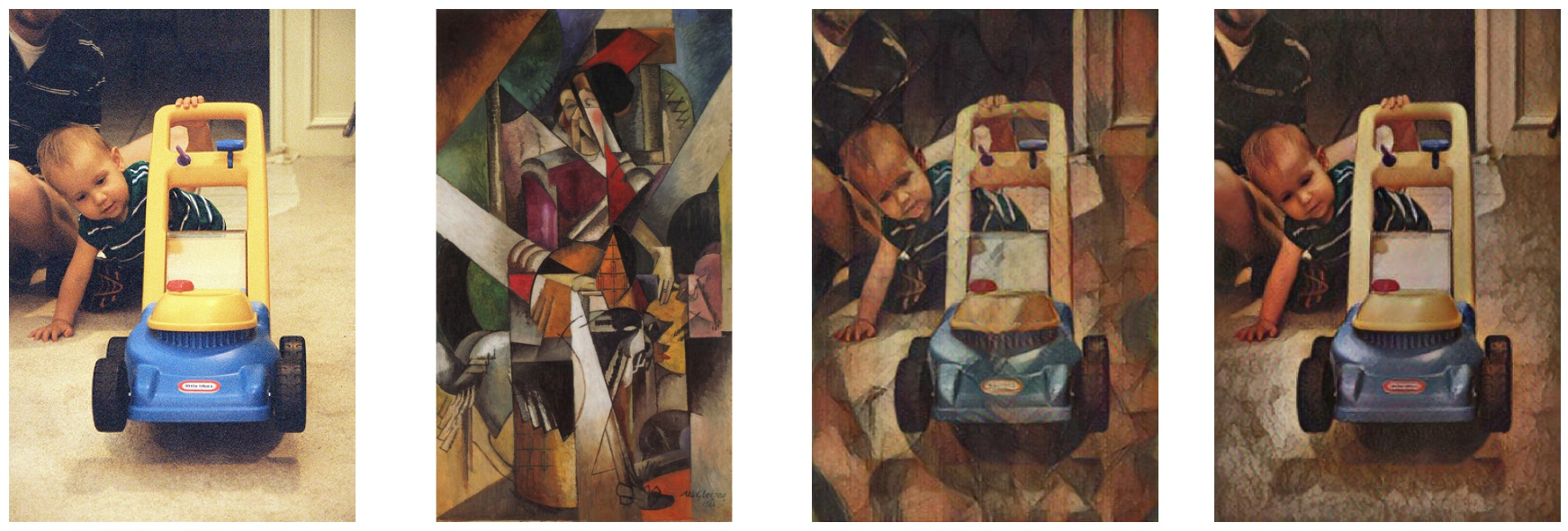}
    \includegraphics[width=\linewidth,trim=0 1cm 0 1cm,clip]{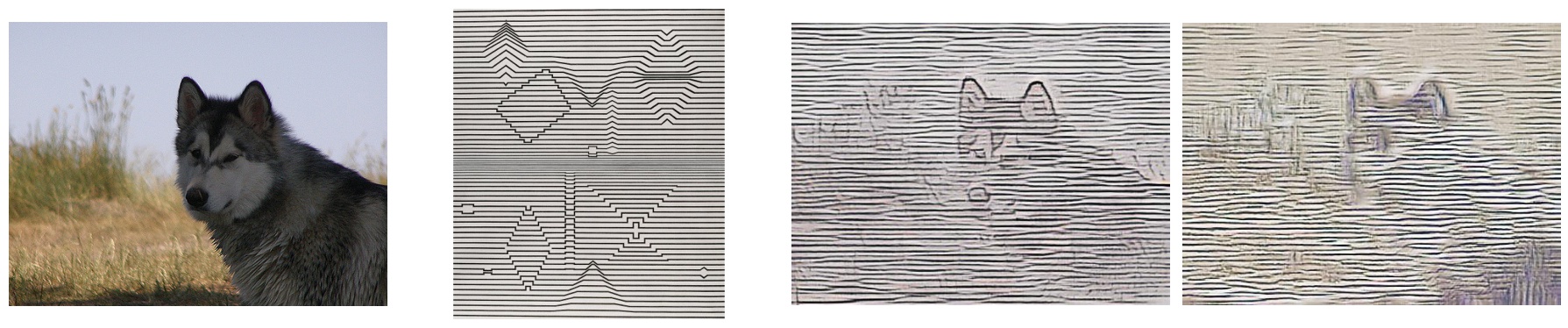}
    \caption{Additional qualitative results for Google Magenta\cite{megenta2}}
    \label{fig.appendix_magenta}
\end{figure*}

\begin{figure*}
    \centering
    \begin{tabular}{p{4cm}p{4cm}p{4cm}p{4cm}}
        \centering{Content} & \centering{Style} & \centering{Our Balanced Style Loss} & \centering{Classic Style Loss}
    \end{tabular}
    
    \includegraphics[width=\linewidth,trim=0 0.5cm 0 0.5cm,clip]{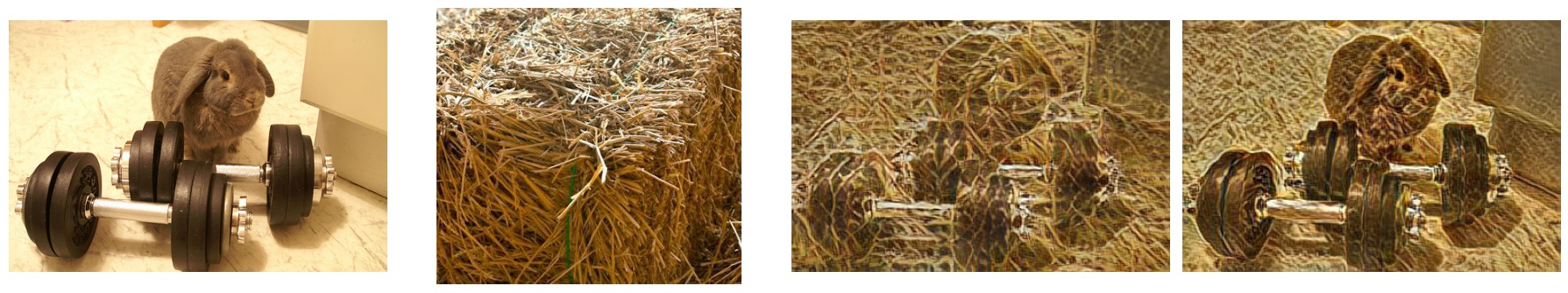}
    \includegraphics[width=\linewidth,trim=0 0.5cm 0 0.5cm,clip]{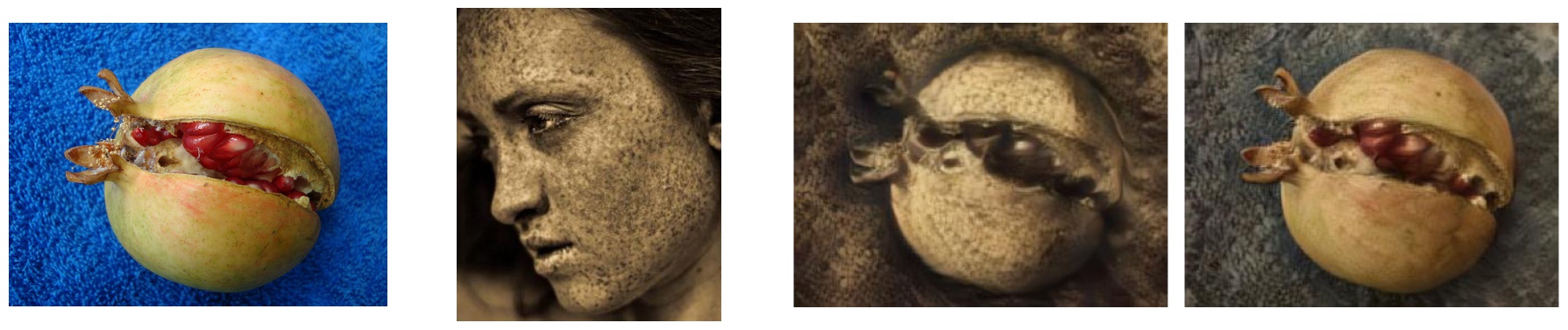}
    \includegraphics[width=\linewidth,trim=0 0.8cm 0 0.8cm,clip]{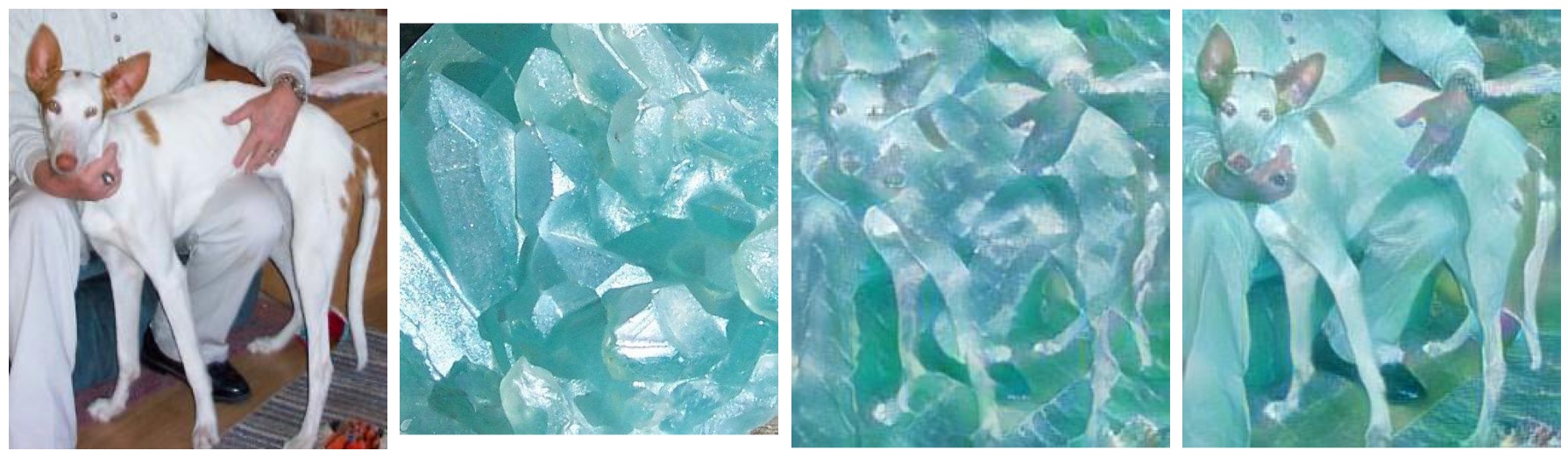}
    \includegraphics[width=\linewidth,trim=0 2cm 0 1cm,clip]{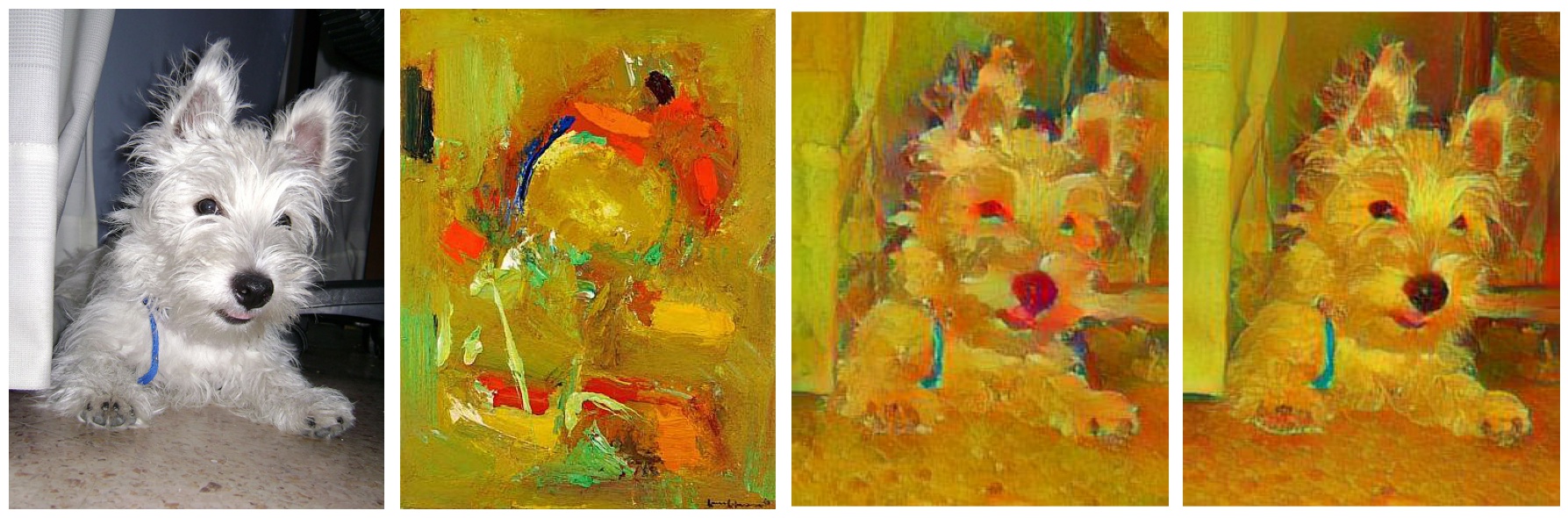}
    \includegraphics[width=\linewidth,trim=0 0.5cm 0 0.5cm,clip]{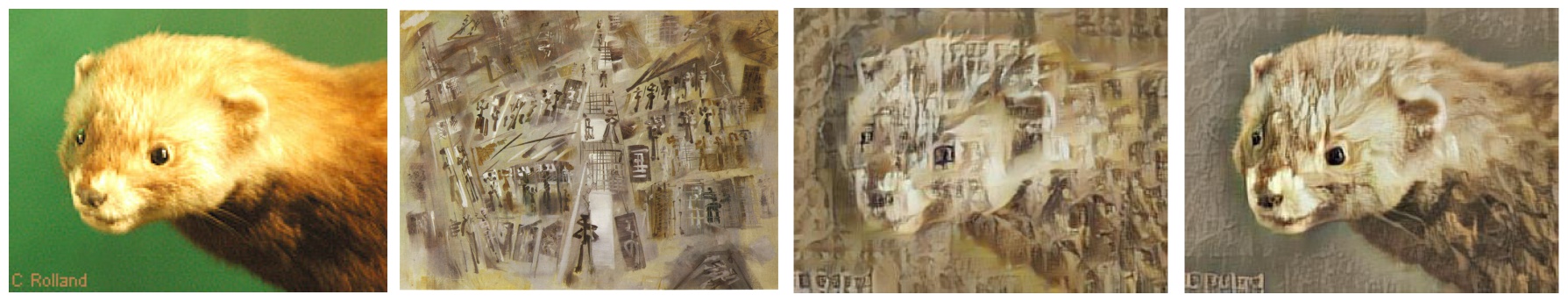}
    \includegraphics[width=\linewidth,trim=0 0.5cm 0 0.5cm,clip]{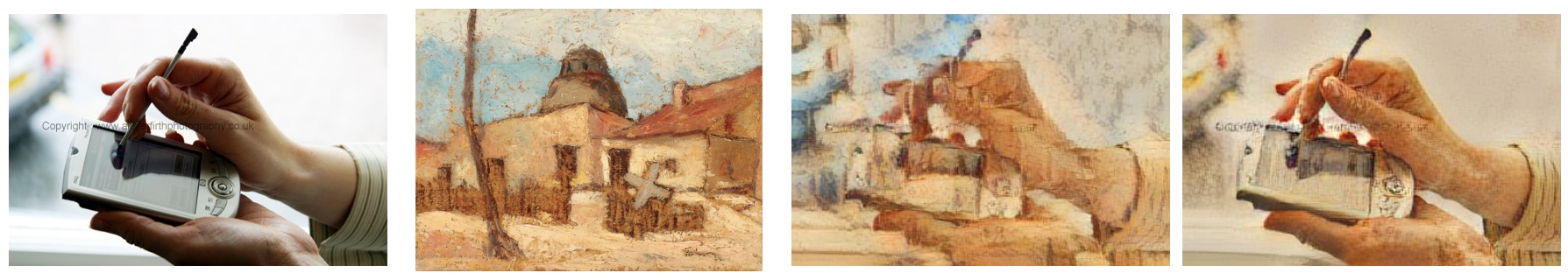}
    \caption{Additional qualitative results for LinearTransfer\cite{LinearTransfer}}
    \label{fig.appendix_LinearTransfer}
\end{figure*}

\begin{figure*}
    \centering
    \begin{tabular}{p{4cm}p{4cm}p{4cm}p{4cm}}
        \centering{Content} & \centering{Style} & \centering{Our Balanced Style Loss} & \centering{Classic Style Loss}
    \end{tabular}
    
    \includegraphics[width=\linewidth,trim=0 1.5cm 0 0.5cm,clip]{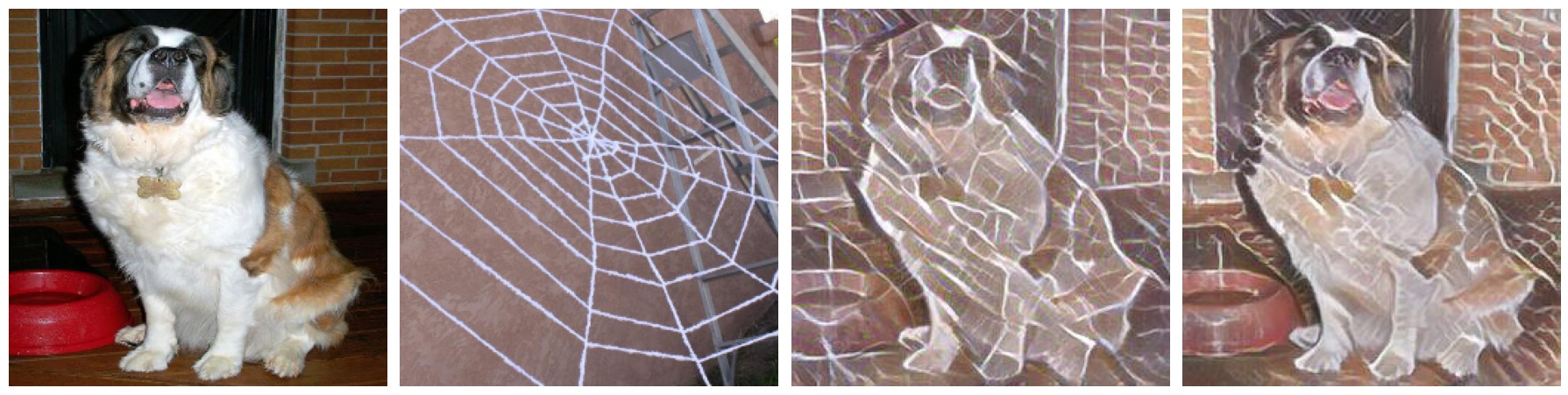}
    \includegraphics[width=\linewidth,trim=0 1cm 0 1cm,clip]{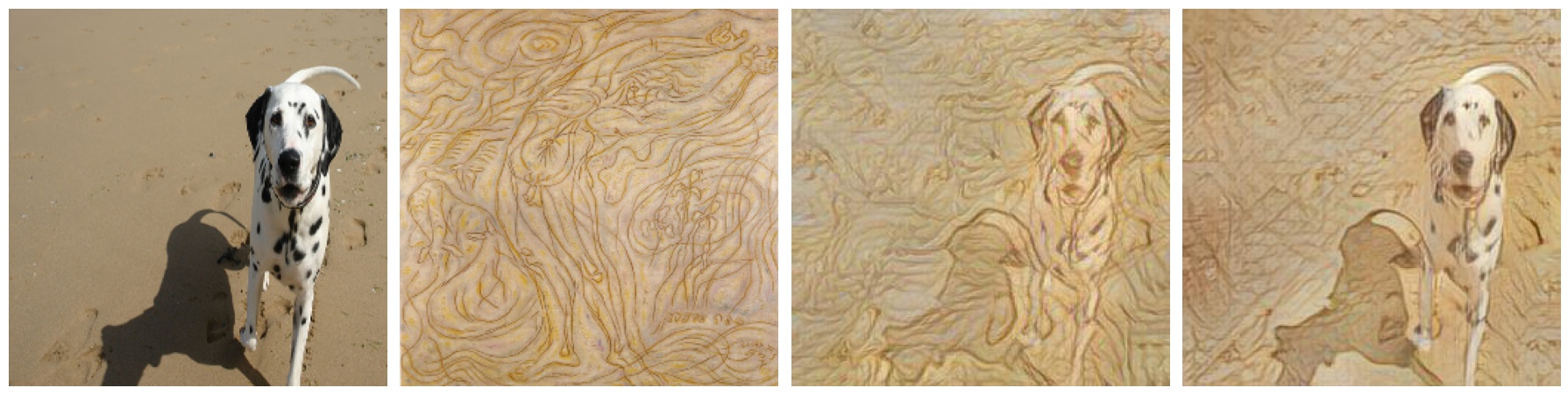}
    \includegraphics[width=\linewidth,trim=0 1cm 0 1cm,clip]{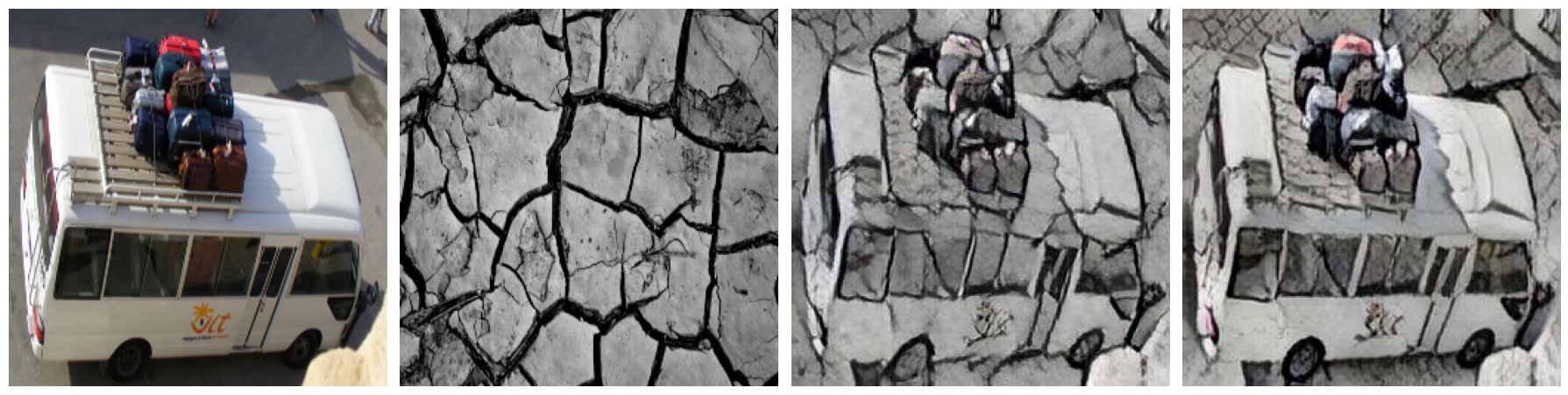}
    \includegraphics[width=\linewidth,trim=0 1cm 0 1cm,clip]{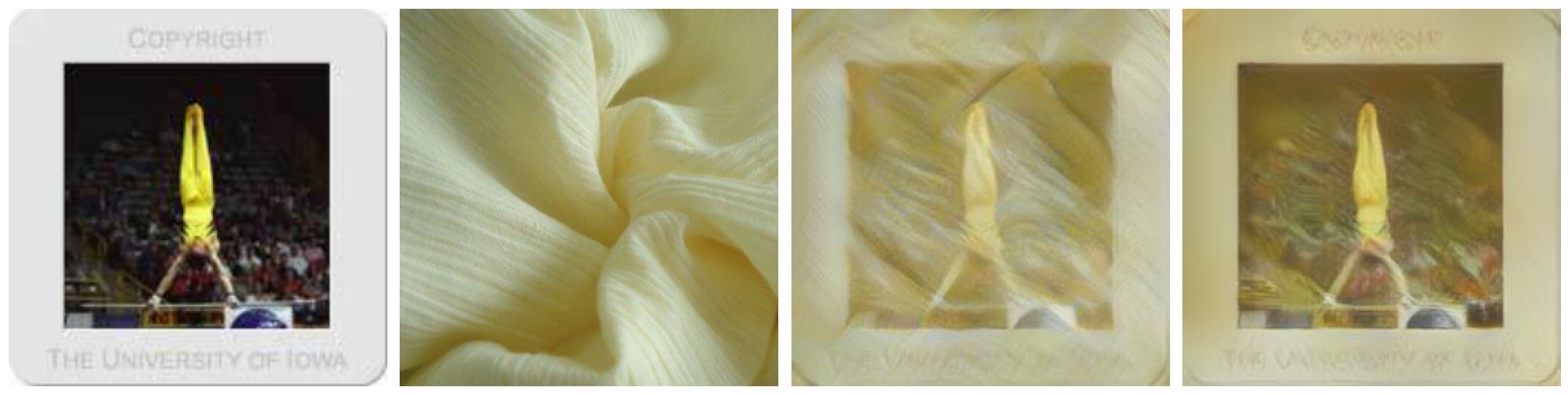}
    \includegraphics[width=\linewidth,trim=0 1cm 0 1cm,clip]{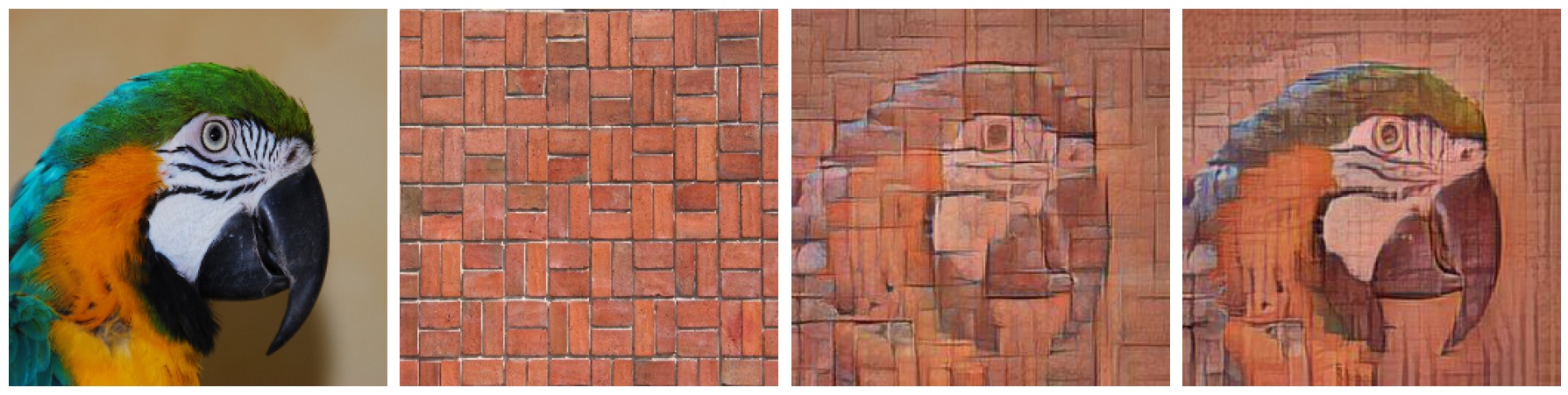}
    \includegraphics[width=\linewidth,trim=0 1cm 0 1cm,clip]{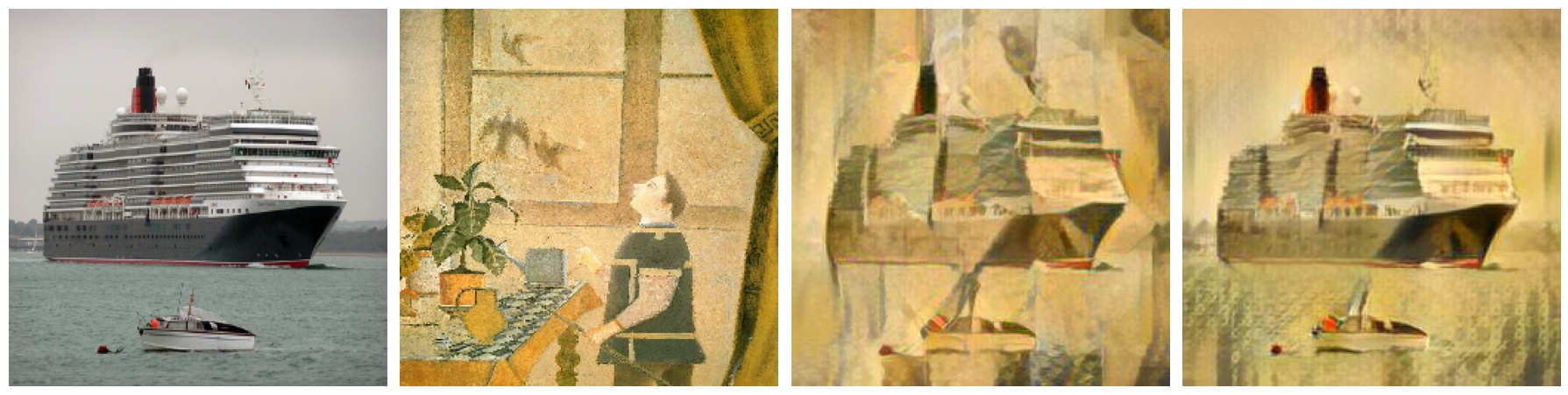}
    \caption{Additional qualitative results for SANet\cite{SANet}}
    \label{fig.appendix_SANET}
\end{figure*}

We show addtional qualitative results with comparison between the classic AST style loss and our balanced loss in \Cref{fig.appendix_adain,fig.appendix_magenta,fig.appendix_LinearTransfer,fig.appendix_SANET}.

\end{document}